\documentclass{article}

\PassOptionsToPackage{numbers,compress}{natbib}

\usepackage{graphics, enumitem}
\newenvironment{denseitemize}{
\begin{itemize}[topsep=2pt, partopsep=0pt, leftmargin=1.5em]
  \setlength{\itemsep}{2pt}
  \setlength{\parskip}{0pt}
  \setlength{\parsep}{0pt}
}{\end{itemize}}

\usepackage[preprint]{neurips_2026}

\usepackage[utf8]{inputenc}
\usepackage[T1]{fontenc}
\usepackage{hyperref}
\usepackage{url}
\usepackage{booktabs}
\usepackage{multirow}
\usepackage{amsfonts}
\usepackage{nicefrac}
\usepackage{microtype}
\usepackage[table]{xcolor}
\usepackage{graphicx}
\usepackage{caption}
\usepackage{subcaption}
\usepackage{float}
\usepackage{tikz}
\usepackage{pgfplots}
\usepackage{algorithm}
\usepackage{algpseudocode}
\usepackage{amsmath, amssymb, bm}
\usepackage{wrapfig}
\usepackage{titlesec}
\usepackage{amssymb}
\newcommand{\R}{\mathbb{R}}
\titlespacing*{\section}
{0pt}    
{1.0ex}  
{0.5ex}  

\titlespacing*{\subsection}
{0pt}
{0.8ex}
{0.4ex}

\titlespacing*{\paragraph}
{0pt}
{0.5ex}
{1em}

\definecolor{algcyan}{RGB}{0,128,160}   
\newcommand{\algcomment}[1]{\State \textcolor{algcyan}{\textit{// #1}}}

\pgfplotsset{compat=1.18}
\usepackage{xspace}

\title{SparseSAM: Structured Sparsification of Activations in Segment Anything Models}
\author{Hoai-Chau Tran$^{1, 3}$,
  Chi H. Nguyen$^{2, 3}$
  Duy M. H. Nguyen$^{4, 5, 6}$, \\ 
  \textbf{Mathias Niepert}$^{5,6}$, 
  \textbf{Fan Lai} $^{1}$, 
  \textbf{Khoa D. Doan}$^{2,3}$. \\
  $^{1}$ University of Illinois at Urbana-Champaign \\
  $^{2}$ College of Engineering \& Computer Science, VinUniversity, \\
  $^{3}$ VinUni-Illinois Smart Health Center, VinUniversity, $^{4}$ DFKI \\
  $^{5}$Max Planck Research School for Intelligent Systems (IMPRS-IS), 
  $^{6}$University of Stuttgart,
  \\
  \texttt{\{chauht2\}@illinois.edu} 
}
\newcommand{\name}{SparseSAM\xspace}

\begin{document}

\maketitle

\begin{abstract}
The Segment Anything Model (SAM) achieves strong open-vocabulary segmentation, but its ViT-based image encoders dominate inference latency and memory. Existing activation compression methods, such as token merging, reduce the token length to process, yet introduce non-trivial runtime overhead and encounter catastrophic quality drop under high compression. Other methods applying Sparse Attention focus on attention alone, leaving the MLP fully dense and capping achievable speedup.  We propose \emph{SparseSAM}, a \textit{(i) training-free structured sparsification} framework that jointly accelerates attention and MLP layers while preserving token identity. SparseSAM introduces \textit{(ii) Stripe-Sort Attention}, which uses a deterministic Z-order permutation to transform dense attention into static hardware-friendly sparse patterns, eliminating dynamic masking overhead. SparseSAM further introduces a \textit{(iii) Residual-Consistency MLP} that routes only informative tokens through the MLP while propagating remaining tokens through the residual pathway. Across four segmentation benchmarks, SparseSAM loses only 0.004 mIoU at a 0.4 density and 0.021 mIoU at 0.3, a 2.10× reduction in accuracy loss versus token merging advances, while achieving 2x faster inference and 2.8× memory reduction.

\end{abstract}

\section{Introduction}

SAMs have demonstrated strong performance in image segmentation and are widely adopted across downstream tasks, driven in large part by their strong zero-shot generalization capabilities~\citep{liang2022expediting, liu2017netslimming, liu2019structuredkd, liu2021groupfisher}.
Unlike large language models (LLMs), SAMs follow a modular architecture consisting of an image encoder, a prompt encoder, and a lightweight mask decoder. The image encoder overwhelmingly dominates both computation and model parameters (by over 99\%). Serving SAM in latency-sensitive or resource-constrained environments, such as large-scale serving systems or edge devices, remains prohibitively expensive.

To meet the pressing demand for efficient inference, recent advances have explored replacing SAM's image encoder with more compact architectures~\citep{zhang2023faster, zhou2023edgesam, xiong2023efficientsam, zhao2023fastsam}. While reducing inference cost, they typically require training an entirely new model, incurring substantial training and data collection overheads, yet often introduce non-trivial accuracy degradation.
An alternative direction focuses on training-free, post-training optimization, such as exploiting attention sparsity through customized kernels originally designed for LLMs and diffusion models \citep{jiang2024minference, xiao2024duoattention, xi2025sparse}. By skipping less important attention blocks, these methods sidestep dense computation but encounter an architectural mismatch due to SAM's multi-scale design: SAM partitions each image into a large number of local regions, producing hundreds of attention heads with fewer than 200 tokens each, and at this granularity, the overhead of computing dynamic sparsity masks frequently offsets any theoretical speedups.

Beyond attention, the MLP blocks contribute substantially to the encoder latency, necessitating holistic compression strategies. Existing efforts, however, remain insufficient. 
MLP pruning advances \citep{chen2024slimsam} require expensive retraining or knowledge distillation. 
Post-training quantization \citep{lv2024ptq4sam, wang2024q, ranjan2025mix, zhang2025ahcptq} targets low-bit representations (e.g., W4A4) but rarely delivers measurable GPU speedups due to SAM's relatively small matrix dimensions. Similarly, activation compression via token merging \citep{bolya2022token, tran2024accelerating, tran2025many} is also ill-suited for this SAM. Because segmentation requires the model to preserve the full token set at the output, every merge operation must be followed by a corresponding unmerge step to restore the original spatial resolution. This per-layer administrative overhead ultimately outweighs the computational savings, frequently resulting in higher overall latency and degraded quality (Fig.~\ref{fig:hq44k}).

\begin{figure}[!htb]
  \centering
  \includegraphics[width=\textwidth]{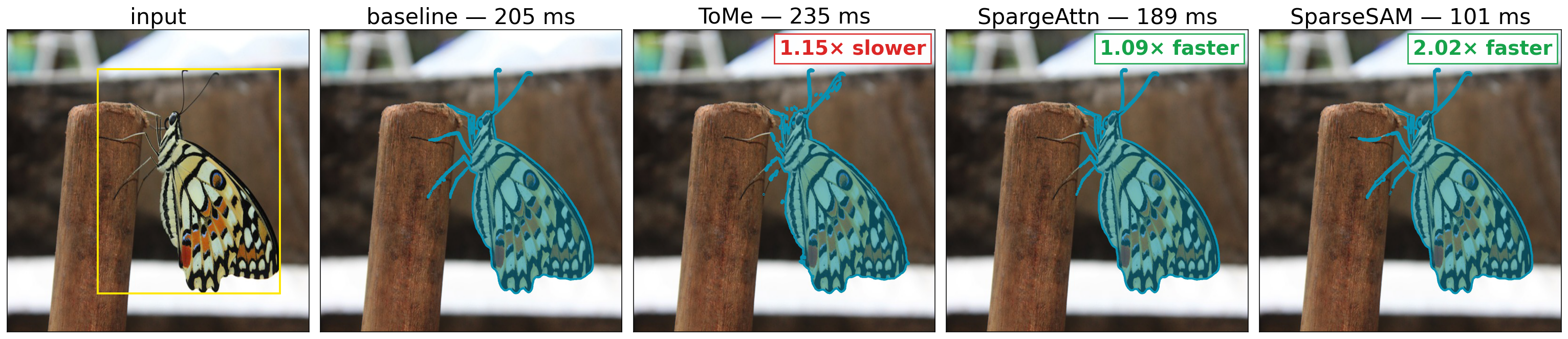}
  \caption{\name enables 2.02$\times$ faster SAM inference while preserving quality.}
  \vspace{-0.15in}
  \label{fig:intro-seg-example}
\end{figure}

\begin{figure}[t]
\centering
\resizebox{1.0\linewidth}{!}{\input{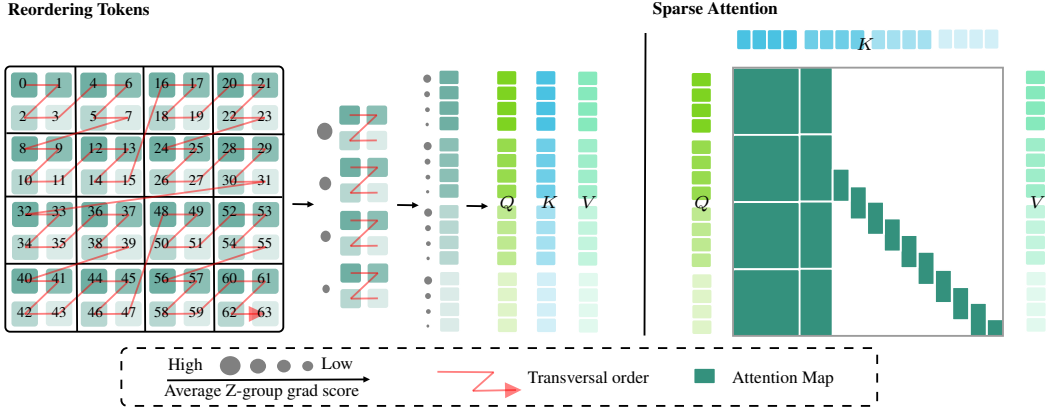}}
\caption{We employ a deterministic \textbf{Z-order (Morton) traversal} to linearize 2D spatial tokens into a 1D sequence. This traversal naturally preserves spatial locality by grouping neighboring pixels into localized "Z-groups." Such grouping enables the efficient computation of gradient-based importance scores within local windows (Eq.~\ref{eq:energy}), which are then used to rank and route tokens. The resulting permuted sequence transforms the attention map into a hardware-friendly block-banded structure, consisting of a dense global prefix for high-saliency tokens and a banded diagonal for local spatial interactions.}
\vspace{-0.15in}
\label{fig:attn-reorder_overview}
\end{figure}

These limitations point to a key gap: \emph{efficient, training-free methods that jointly reduce computation in both attention and MLP layers, while preserving token-level fidelity required for segmentation.} 
We introduce \emph{SparseSAM}, a training-free approach designed to leverage the inherent sparsity exposed by the activations in both the attention and MLP layers of SAM. We propose two complementary techniques that improve encoder throughput while preserving segmentation quality.
\begin{denseitemize}
    \item \textbf{Stripe-Sort Attention.} We introduce a new structured sparse attention mechanism based on a deterministic Z-order (Morton) permutation~\cite{morton1966computer,erkan2023space,stucker2026jz}. This reordering induces spatial locality in token indices, enabling a static, block-structured sparsity pattern implemented via a custom CUDA kernel. The resulting pattern consists of a banded diagonal capturing local interactions, augmented with a small global keep set for long-range dependencies. Notably, this design avoids dynamic mask construction and preserves the full key/value set, ensuring both efficiency and quality (Fig.~\ref{fig:attn-reorder_overview} and Fig.~\ref{fig:perm_example}).
    \vspace{0.05in}
    \item \textbf{Residual-Consistency MLP.} We propose a token routing mechanism that applies the MLP only to a top-$K$ subset of tokens selected based on their importance (i.e., the high-rank prefix of the sorted z-group tokens), while allowing the remaining tokens to bypass the MLP through the residual connection. This design retains information flow without incurring the full cost of dense MLP computation (Fig.~\ref{fig:attn-reorder}).
\end{denseitemize}
Both techniques utilize a \textbf{static, one-shot permutation set} that is computed once and reused across all model layers, ensuring that per-layer overhead remains negligible. Our evaluations across five distinct SAM checkpoints and five benchmarks demonstrate that SparseSAM achieves an average ~2.0$\times$ inference speedup. Even at high-density compression rates, the method maintains 50\% sparsity with a minimal segmentation accuracy drop ( <1\% IoU loss), as illustrated in Figure \ref{fig:hq44k}.

\section{Background and Related Work }
\label{sec:related}


\begin{wrapfigure}{r}{0.65\textwidth}
\vspace{-.3cm}
    \centering
    \includegraphics[width=\linewidth]{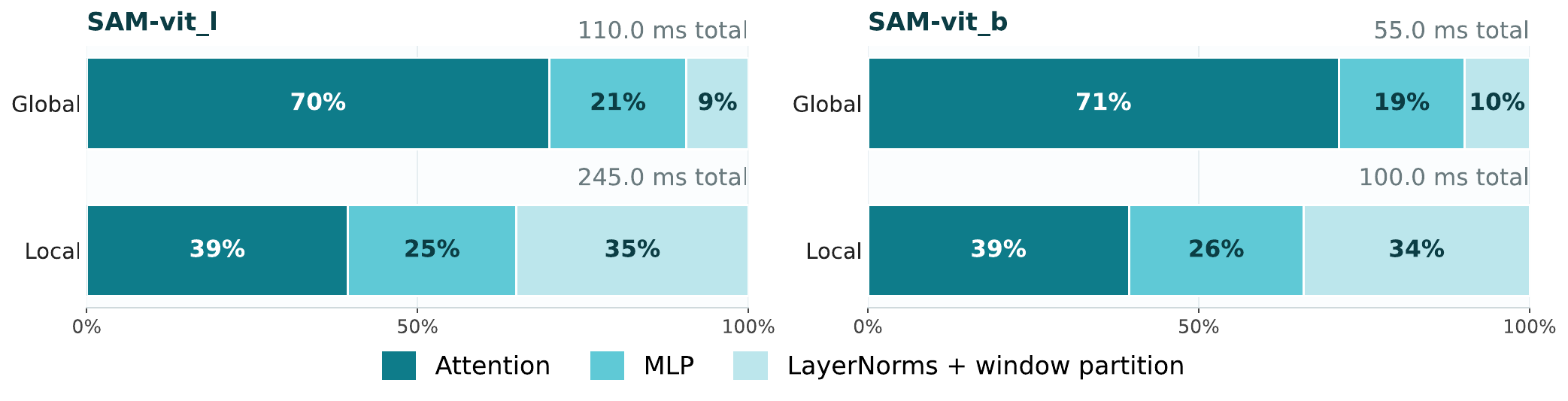}
    \caption{SAM latency profiling. }
    \label{fig:profiling}
    \vspace{-.3cm}
\end{wrapfigure}

\textbf{SAM Architecture.} SAM adopts a hierarchical Vision Transformer (ViT) encoder composed of both local and global attention blocks. Local blocks partition the image into non-overlapping windows and apply self-attention independently within each window, while global blocks perform dense attention over the entire token sequence to capture long-range dependencies. 


As shown in Fig.~\ref{fig:profiling}, the primary computational bottleneck in global attention blocks is the attention operator, which can account for up to 70\% of runtime. In contrast, \textit{local blocks} exhibit a \textit{dual bottleneck, where both attention and MLP layers} contribute significantly to latency. This distinction is important because SAM architectures contain substantially more local blocks than global blocks. For example, SAM-L contains 20 local blocks but only 4 global blocks. Consequently, optimizing attention alone provides limited end-to-end acceleration; practical speedups require jointly reducing the cost of both attention and MLP computation.


\paragraph{Sparse Attention Approaches.}

Recent advancements have leveraged sparsity to mitigate the quadratic cost of Transformer attention. SpargeAttention \citep{zhang2025spargeattention} employs a two-stage online filtering mechanism, while PISA \citep{li2026pisa} utilizes block-wise Taylor expansions to approximate non-critical regions. Additionally, the Sparse Video Gen series \citep{xi2025sparse, yang2025sparse} introduces semantic-aware permutations to cluster salient tokens.

While effective for long-context Transformers and diffusion models, where \textit{sequence lengths reach tens of thousands}, these approaches \textit{encounter a "complexity wall"} when applied to the SAM due to its architecture, which primarily relies on local-window attention with fewer than 200 tokens per head. In this regime, the computational overhead of dynamic mask construction and online importance estimation often outweighs the savings from skipped operations. Furthermore, existing sparse attention methods fail to address the substantial MLP costs incurred by local blocks, thereby capping potential end-to-end acceleration (Fig.~\ref{fig:hq44k}). Unlike these dynamic approaches, our method leverages a \textbf{deterministic Z-order permutation} to induce a static, hardware-efficient sparsity pattern that accelerates both attention and MLP components with zero runtime search overhead.

\paragraph{Activation Compression Approaches.}


A complementary research direction focuses on reducing activation density to jointly accelerate attention and MLP layers. Weight-activation quantization \citep{lv2024ptq4sam, wang2024q, ranjan2025mix, zhang2025ahcptq} reduces theoretical FLOPs and memory footprints; however, modern INT4 Tensor Core kernels require large contraction dimensions to saturate GPU throughput~\cite{kim2023marlin}. Since SAM operates on relatively small matrix tiles (typically $64$ to $1024$), the fixed overheads of quantization, including scaling factor computation and kernel launch latency; thus often negate the gains from reduced precision.

Alternatively, token merging \citep{bolya2022token, tran2024accelerating, tran2025many} reduces the effective sequence length by coalescing redundant tokens. While successful in global classification tasks, this approach is fundamentally ill-suited for high-fidelity segmentation, where the decoder demands full spatial resolution. Restoring this resolution necessitates frequent per-layer merge-unmerge cycles, introducing scatter-gather overheads \citep{nguyen2026structsam} that typically exceed the computational savings. Furthermore, aggressive merging erodes fine-grained token distinctiveness, leading to a precipitous decline in mask quality at higher compression rates. In contrast, our approach utilizes a \textbf{structure-preserving} Z-order sparsity that retains the original token grid, bypassing both the latency of quantization and the information loss of merging.

\section{Observations and Motivation}
\begin{wrapfigure}{r}{0.40\textwidth}
\vspace{-.3cm}
    \centering
    \vspace{-0.3in}
    \includegraphics[width=0.35\textwidth]{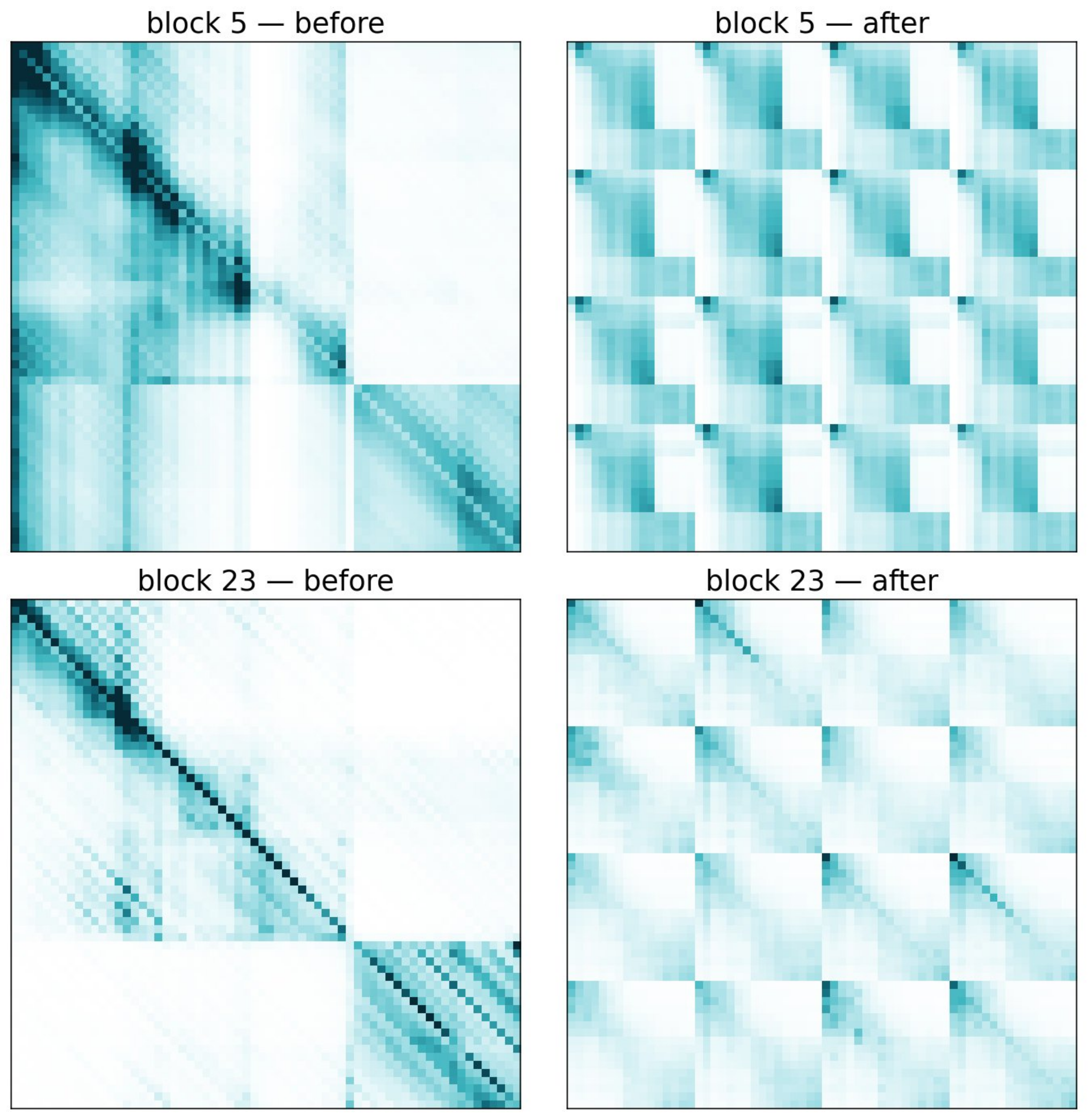}
    \caption{\small{\textbf{Permuted attention visualization}. Comparison between the original SAM attention pattern (left) and our Z-order permuted pattern (right). The permutation induces a banded diagonal structure, allowing for significant hardware acceleration with minimal loss in spatial information.}}
    \label{fig:attn_perm}
    \vspace{-0.5in}
\end{wrapfigure}
Since SAM is exceptionally sensitive to overheads, we require an approach that minimizes per-layer administrative costs while maintaining high-fidelity feature representations.

Rather than dynamically identifying sparse regions, we explore an alternative perspective: \emph{can we restructure the token layout itself such that efficient sparsity patterns emerge deterministically?} We next introduce the key insights that motivate our work. 

\vspace{-0.05in}
\paragraph{Observation 1: Structured token permutation induces reusable sparse attention patterns.}
Our design is motivated by prior activation compression studies~\citep{tran2024accelerating, tran2025many}, which show that preserving spatial diversity during token compression is substantially more important than preserving contiguous foreground regions. Intuitively, segmentation quality depends heavily on maintaining broad spatial coverage across the image. Therefore, we seek a way to permute the $\mathbf{Q}$ and $\mathbf{K}$ matrices such that the resulting attention computation effectively performs spatial downsampling while preserving the global distribution of tokens within each sub-attention map. 

To this end, we introduce the \textbf{stripe-sort attention kernel}, which applies a parameter-free, deterministic permutation. This reordering incurs minimal overhead and can be readily fused into existing attention kernels. 
As illustrated in Figure~\ref{fig:attn_perm}, the permutation reorganizes tokens such that the original large attention map is decomposed into four smaller sub-attention maps, each exhibiting a similar attention pattern. In effect, the operation produces multiple phase-shifted views of the full attention, while maintaining broad spatial coverage. Further details are provided in Section~\ref{sec:method}.

\begin{wrapfigure}{r}{0.45\textwidth}
    \centering
    \includegraphics[width=0.43\textwidth]{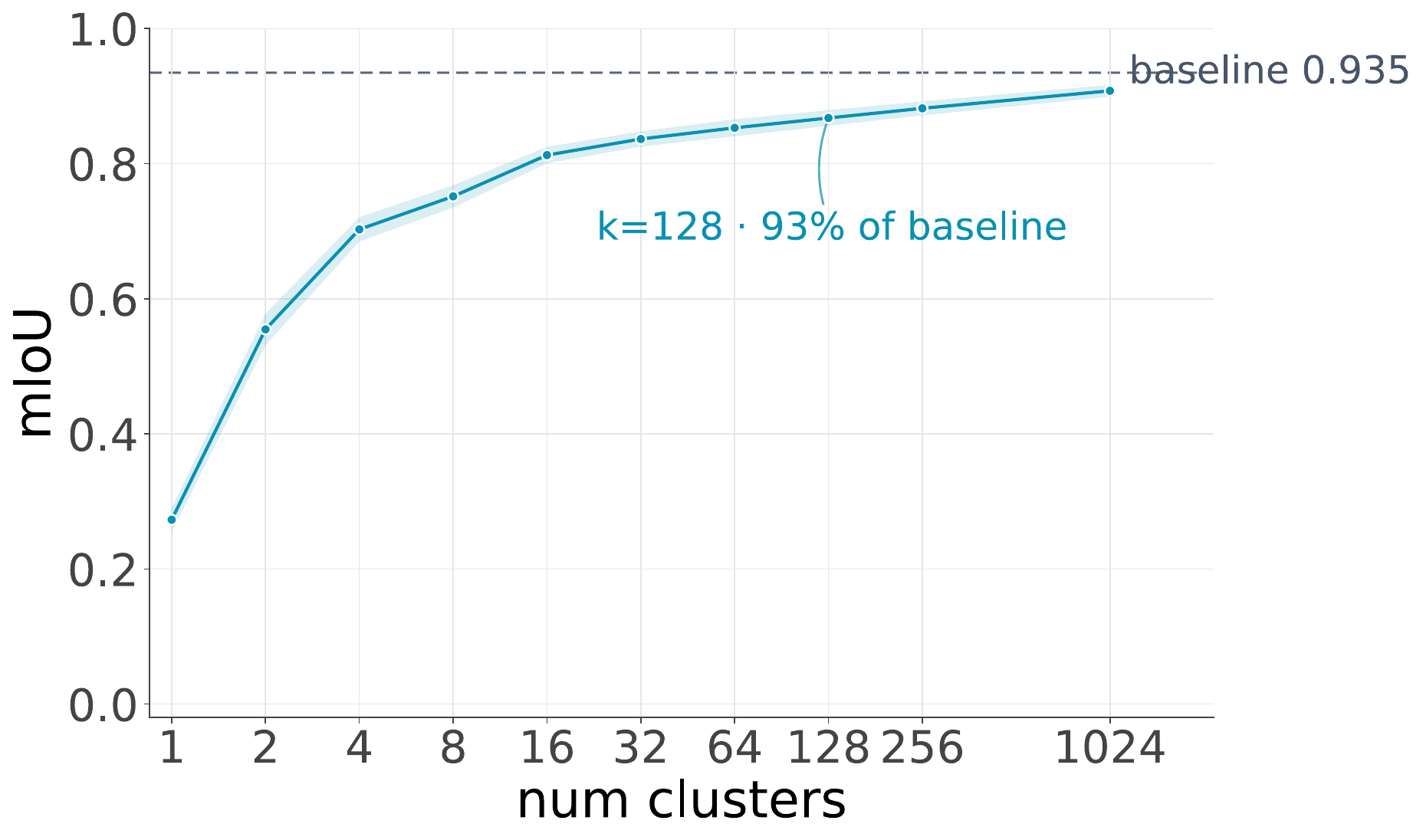}
    \caption{K-means clustering replacement.}
    \label{fig:cluster-probe}
    \vspace{0.5em}
    \centering
    \small
    \setlength{\tabcolsep}{4pt}
    \begin{tabular}{l|cccc}
        \toprule
        & Layer 1 & Layer 6 & Layer 16 & Layer 22 \\
        \midrule
        $\rho$ & 0.64 & 0.62 & 0.75 & 0.78 \\
        \bottomrule
    \end{tabular}
    \captionof{table}{Layer-wise correlation  between token dissimilarity and the norm of $||\Delta_{MLP}||^2$ for each tokens.}
    \label{tab:cluster-sim}
     \vspace{-.3cm}
\end{wrapfigure}
\paragraph{Observation 2: SAM's decoder depends more on inter-region contrast than precise per-token representations.}
We further observe that SAM exhibits substantial representational redundancy, stemming from how the mask decoder utilizes encoder outputs. To investigate this, at each block we perform $k$-means clustering on the attention outputs and replace the encoder’s $4096$ tokens after the MLP update with their nearest cluster centroids. Remarkably, even with $k = 128$ (\textit{approximately $3\%$ of the original token counts}), the decoder retains \emph{93\% of the baseline mIoU} on COIFT (Fig.~\ref{fig:cluster-probe}), with performance saturating near the baseline at $k = 256$. These results suggest that tokens within the same cluster are largely interchangeable from the decoder’s perspective. What matters is the \emph{contrast between clusters}, which enables the model to distinguish different objects, rather than the precise representation of each individual patch.

\paragraph{Observation 3: Which tokens actually require expensive MLP updates?}  We analyze how MLP layers modify token representations and find that SAM’s MLP exhibits a \textit{strong inductive bias toward separating distinct image regions}: a large fraction of background tokens remain close to the residual branch, while semantically rich foreground tokens undergo significantly larger updates (Fig.~\ref{fig:mlp-routing}). To better quantify this behavior, in Table~\ref{tab:cluster-sim} we measure the correlation between the MLP update magnitude, $||\Delta_{\text{MLP}}||_2$, and token dissimilarity (measured via cosine distance). We observe that tokens that are more distinct in feature space tend to receive larger updates, with a strong positive correlation ($\rho \approx 0.6\text{–}0.7$).

This redundancy motivates our second contribution, \emph{residual-consistency MLP}, which prioritizes MLP updates for semantically important tokens while maintaining lightweight residual propagation for the remaining tokens. 

\section{Methods}
\label{sec:method}

\subsection{Stripe-Sort Attention}
\label{subsec:attention}

We first introduce \emph{Stripe-Sort Attention}, a structured sparse attention mechanism that reorganizes token layouts to expose deterministic sparsity patterns amenable to efficient GPU execution. Specifically, we use a deterministic, parameter-free permutation that decomposes the $N \times N$ attention matrix into a $4 \times 4$ mosaic of $(N/4) \times (N/4)$ sub-maps. Since the permutation depends only on spatial position, it can be precomputed once and statically compiled into the attention kernel with no runtime overhead. Importantly, the four resulting token subsets are phase-shifted, half-resolution views of the original image, where each subset remains spatially distributed across the entire input rather than localized to a single region. Consequently, each sub-map preserves global context while operating on only a quarter of the tokens.

\paragraph{Gradient-Based Ordering.}
Given an input feature map $\mathbf{X} \in \mathbb{R}^{H \times W \times D}$, our goal is to construct a one-dimensional token ordering that places high-information regions early in the sequence while spreading them evenly across spatial positions. The first property ensures that when the sequence is later truncated or sparsified, the most informative tokens are retained, while the second property ensures that the retained tokens collectively cover the full spatial extent of the image rather than concentrating in a single region. 

To achieve both properties, we use local image gradients as a lightweight proxy for information density. Intuitively, regions with large gradient magnitude correspond to object boundaries, texture, and visually salient structures, while regions with low gradient magnitude correspond to smooth or homogeneous areas that contribute relatively little to dense prediction.
\begin{wrapfigure}{r}{0.375\textwidth}
  \centering
  \includegraphics[width=\linewidth]{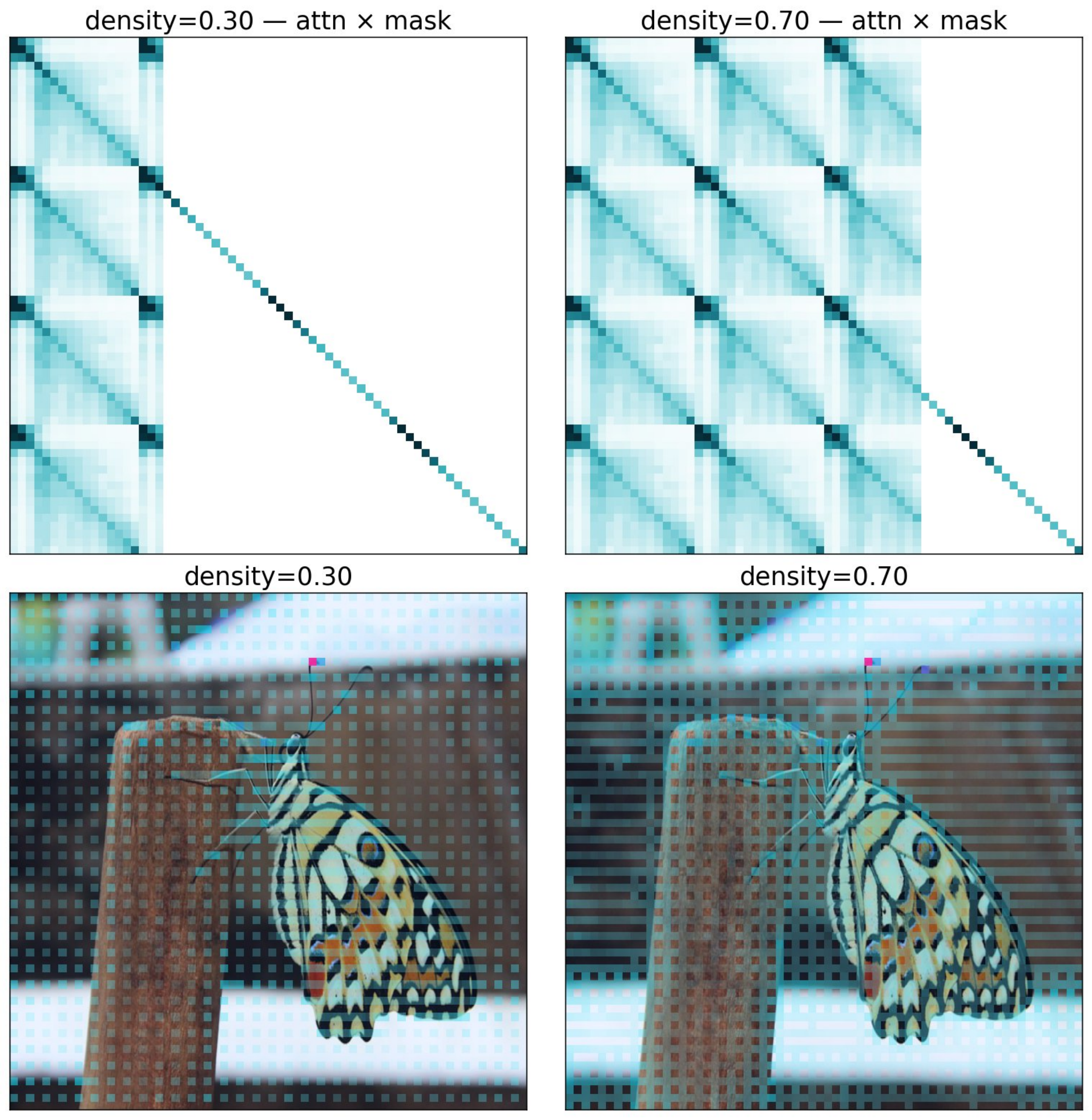}
  \caption{Effect of Stripe-Sort Attention with different sparsity ratios}
  \label{fig:perm_example}
\end{wrapfigure}
We measure local gradient magnitude using the Sobel operator \cite{gonzalez2009digital}, a classical edge-detection filter that approximates the spatial derivatives of an image through a pair of fixed $3 \times 3$ convolutional kernels. Specifically, the gradient magnitude at each spatial position is
\begin{equation}
\mathbf{M}[i,j] = \sqrt{(\mathbf{S}_x * \mathbf{X})^2[i,j] + (\mathbf{S}_y * \mathbf{X})^2[i,j]},
\label{eq:energy}
\end{equation}
where $\mathbf{S}_x$ and $\mathbf{S}_y$ denote the horizontal and vertical Sobel kernels, and $*$ represents channel-wise 2D convolution followed by summation across the $D$ channels. The resulting map $\mathbf{M} \in \mathbb{R}^{H \times W}$ is a single-channel saliency map, where high $ \mathbf{M} $ values correspond to object boundaries and low values correspond to smooth regions. This computation is fully deterministic, parameter-free, and only performed once in the first SAM layer, introducing negligible overhead. Sorting the $N = HW$ spatial positions in descending order of $\mathbf{M}$ produces a permutation $\pi$, placing high-gradient regions at the front and low-gradient regions at the back.

\paragraph{Stripe-Sort Permutation.} 
To prevent foreground regions from gathering at the start of the token set, we view $\pi$ as a matrix $\mathbf{T} \in \mathbb{N}^{(N/G) \times G}$ with $\mathbf{T}[t, g] = \pi[t \cdot G + g]$ and define the final scan order as $\sigma = \operatorname{flatten}\bigl(\mathbf{T}^\top\bigr).
$

This visits every $G$-th element of $\pi$ before returning to the next offset, so each of the $G$ resulting blocks of $\sigma$ contains a uniformly subsampled mixture of tokens drawn from across the entire image. As demonstrated in Figure~\ref {fig:perm_example}, applying this reordering to both queries and keys transforms global attention into a deterministic block-diagonal pattern that can be efficiently executed using static block-sparse kernels like FlashAttention. Our method avoids runtime mask generation, semantic clustering, and online scheduling overhead.

We implement Stripe-Sort Attention as a fully fused CUDA kernel to eliminate intermediate memory scattering and redundant kernel launches. In local layers, each sub-image contains at most $196$ tokens, making standard FlashAttention~\citep{dao2023flashattention, dao2023flashattention2} and FlashInfer~\citep{ye2025flashinfer} kernels inefficient due to their fixed tile sizes (e.g., $128 \times 128$). We therefore design a custom CUDA kernel with flexible tiling, using $32 \times 32$ tiles for local layers and $128 \times 128$ tiles for global layers, while maintaining efficient \texttt{mma.sync.aligned.m16n16k16} tensor-core execution. The pseudo-code of the kernel is provided in Appendix \ref{sec:flash_kernel}.


\begin{figure}[t]
    \centering
    \vspace{-0.1in}
    \begin{subfigure}[t]{0.55\textwidth}
        \centering
        \resizebox{0.8\linewidth}{!}{\input{images/res_mlp}}
        \caption{Residual-consistency MLP.}
        \label{fig:attn-reorder}
    \end{subfigure}
    \hfill
    \begin{subfigure}[t]{0.425\textwidth}
        \centering
        \includegraphics[width=0.8\linewidth]{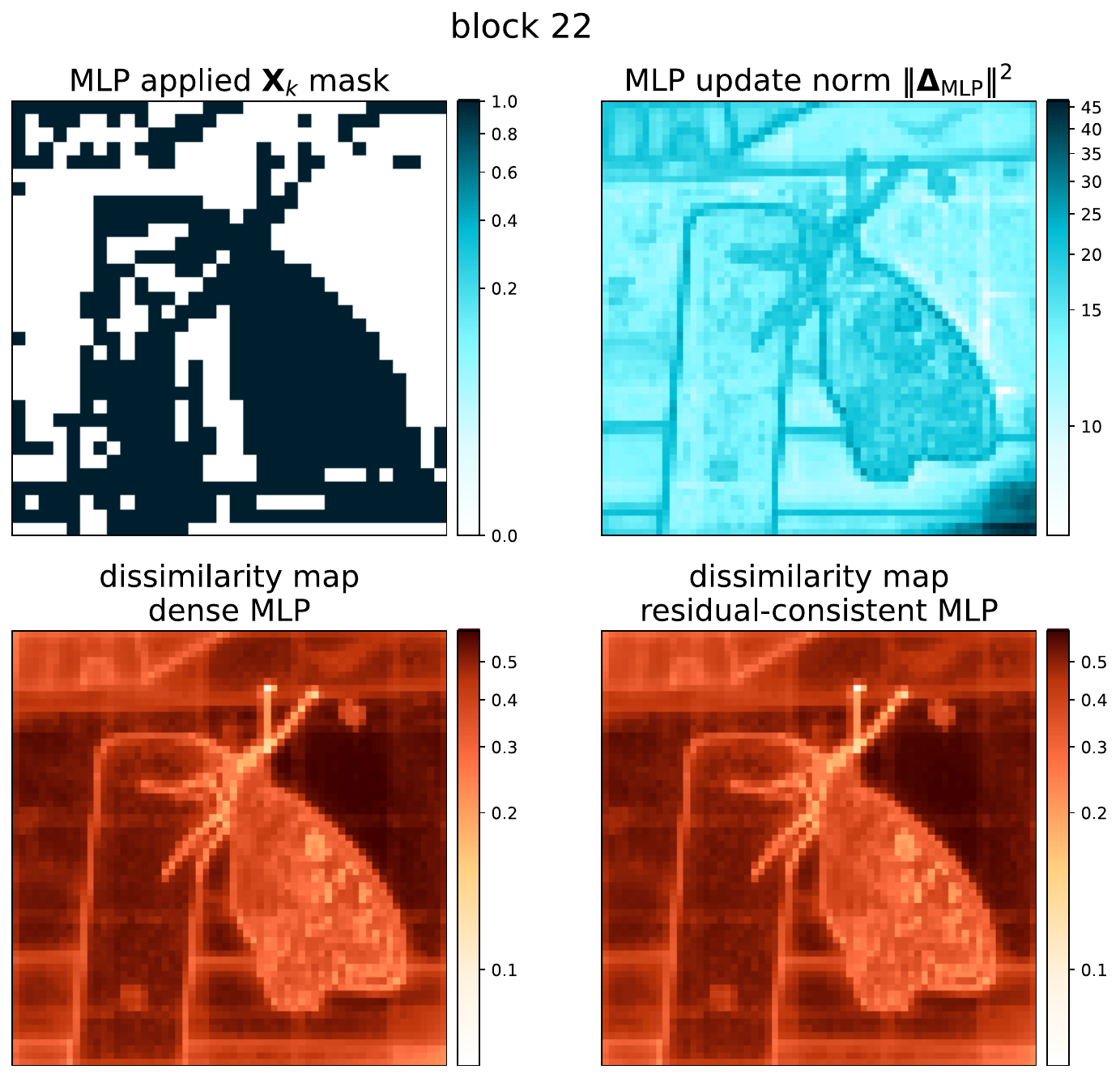}
        \caption{Token keep-set and MLP routing.}
        \label{fig:mlp-routing}
    \end{subfigure}
    \caption{Visualization of the residual-consistency MLP. (a) The deterministic token order $\pi$ is reused to partition tokens into a keep-set $\mathbf{X}_k$, updated by the MLP, and a residual set $\mathbf{X}_r$, which bypasses the MLP unchanged. (b) Our analysis shows that dense SAM implicitly focuses MLP updates on a small set of representative, highly dissimilar tokens. After applying the residual-consistency MLP, the inter-token similarity structure remains close to the dense baseline, preserving the representational geometry required by the segmentation decoder.}
    \vspace{-0.15in}
    \label{fig:method-overview}
\end{figure}

\subsection{Residual-Consistency MLP}
\label{sec:residual-mlp}
\vspace{-0.05in}
Our second contribution is motivated by the structure of MLP updates: the decoder depends more on inter-region contrast than precise per-token representations. Given input tokens $\mathbf{X}=[\mathbf{x}_1,\ldots,\mathbf{x}_N]\in\mathbb{R}^{N\times d}$, the MLP update for token $i$ is $
\Delta_i = \mathrm{MLP}(\mathrm{LN}(\mathbf{x}_i))$, where $\mathrm{LN}(\cdot)$ denotes LayerNorm. We use the update magnitude $u_i=\|\Delta_i\|_2$ to quantify how strongly each token is modified. 

As illustrated in Fig.~\ref{fig:mlp-routing}, the distribution of $\{u_i\}_{i=1}^N$ is highly skewed across encoder blocks: only a small subset of tokens receives large updates, while most remain close to the residual connection. Spatially, these high-update tokens correspond to regions with strong feature dissimilarity, such as object boundaries and salient textures, whereas smooth regions receive minimal updates. Consistent with this observation, Table~\ref{tab:cluster-sim} shows a strong correlation between token uniqueness and update magnitude $u_i$, suggesting that the dense MLP naturally performs an implicit form of routing.
\[
\mathbf{y}_i =
\begin{cases}
\mathbf{x}_i + \Delta_i, & i\in\mathcal{K}, \\
\mathbf{x}_i + \epsilon_i, & i\notin\mathcal{K},
\end{cases}
\]
where $\mathcal{K}$ denotes a small subset of representative tokens that receive meaningful updates and $\epsilon_i$ is typically much smaller than $\Delta_i$.

\paragraph{Formulation.}
Motivated by this observation, we make the routing behavior explicit. Given we reuse the deterministic permutation order to reorder the tokens $\pi(\mathbf{X})$ and partition the sequence into a keep-set $\mathbf{X}_k=\{\mathbf{x}_i \mid i\in\mathcal{K}\}$ and a residual set $\mathbf{X}_r=\{\mathbf{x}_i \mid i\notin\mathcal{K}\}$. As shown in Fig.~\ref{fig:mlp-routing}(a), the MLP is applied only to the keep-set, while residual tokens bypass the MLP unchanged.
\[
\mathbf{Y}_k = \mathbf{X}_k + \mathrm{MLP}(\mathrm{LN}(\mathbf{X}_k)), \quad \mathbf{Y}_r = \mathrm{LN}(\mathbf{X}_r).
\]

By routing only representative tokens through the MLP, our method preserves the feature geometry of the dense baseline while substantially reducing MLP computation. Despite its simplicity, this approach proves highly effective. As shown in Fig.~\ref{fig:mlp-routing}, the inter-token dissimilarity maps produced by the dense MLP and our residual-consistency MLP are nearly identical, indicating that the routed computation preserves the representational structure required by the segmentation decoder. 

\section{Experiments}
\label{sec:eval}

\subsection{Experimental Setup}
\paragraph{Models and Datasets.}
We implement \name\ on top of the official checkpoints of SAM-B, SAM-L, and SAM-H. Unless otherwise specified, all experiments and latency measurements are conducted on a single NVIDIA A100 GPU. We use HQ-44K~\cite{ke2023segment} to evaluate high-quality fine-grained segmentation and MS-COCO~\cite{lin2014microsoft} to evaluate zero-shot generalization on common object categories. For MS-COCO, we utilize DINO~\cite{zhang2022dino}, H-DETR~\cite{jia2022detrs}, and YOLOX~\cite{ge2021yolox} as the primary object detectors. We compare \name against two categories of baselines:

\paragraph{Attention Sparsification Baselines.} We compare against (i) \textsc{SpargeAttention}~\citep{zhang2025spargeattention} and \textsc{Piecewise Sparse Attention} (PISA)~\citep{li2026pisa}, two state-of-the-art training-free sparse attention frameworks originally designed for large Transformer models. These methods dynamically identify important attention regions to reduce computation while approximately preserving dense attention behavior. To ensure fair comparison on SAM, we extend both implementations with fused relative positional encoding support, which is required by SAM's ViT encoder but absent from their original implementations. 

\paragraph{Activation Compression Methods Baselines.} For both Attention and MLP compression, we compare against Token Merging (ToMe) \citep{bolya2022token} and its gradient-aware variant, StructSAM \cite{nguyen2026structsam}, designed for the SAM model. These methods represent the standard paradigm of using token reduction via feature similarity or gradient matching to lower activation density in Vision Transformers. 

\paragraph{Metrics.}
We report segmentation quality using mean Intersection-over-Union (mIoU) and Boundary IoU (BIoU), which together capture both region-level accuracy and boundary fidelity. For system efficiency, we measure end-to-end latency, peak GPU memory usage, and throughput under varying density (i.e., $100\% - sparsity$) ratios. Unless otherwise specified, all latency results are averaged over multiple runs with synchronized CUDA timing.


\subsection{Segmentation Results}
\subsubsection{Common Objects Segmentation}
\begin{table}[t]
\centering
\definecolor{sparsegreen}{RGB}{28,155,138}
\definecolor{finetunegray}{gray}{0.92}
\newcommand{\attnmarker}{\textcolor{black}{\ensuremath{\circ}}}
\newcommand{\mlpmarker}{\textcolor{sparsegreen}{\ensuremath{\bullet}}}
\caption{Zero-shot segmentation on MS-COCO datasets using DINO as a bounding box detector generates a box prompt for the SAM model. Markers indicate compression type: \protect\attnmarker\ attention-only and \protect\mlpmarker\ attention+MLP.}
\vspace{0.05in}
\scriptsize
\setlength{\tabcolsep}{4pt}
\renewcommand{\arraystretch}{1.12}
\resizebox{0.9\textwidth}{!}{%
\begin{tabular}{l|ccc|ccc|ccc}
\toprule
\multirow{2}{*}{Method} & \multicolumn{3}{c|}{\textbf{SAM-B}} & \multicolumn{3}{c|}{\textbf{SAM-L}} & \multicolumn{3}{c}{\textbf{SAM-H}} \\
& mAP & Density & Speedup & mAP & Density & Speedup & mAP & Density & Speedup \\
\midrule
Base & 0.468 & 100\% & $\times$ 1.00 & 0.495 & 100\% & $\times$ 1.00 & 0.499 & 100\% & $\times$ 1.00 \\
\cmidrule(lr){1-10}
\multirow{2}{*}{\attnmarker\ Sparge Attention}
& 0.447 & 25\% & $\times$ 1.25 & 0.461 & 25\% & $\times$ 1.23 & -- & 25\% & -- \\
& 0.463 & 50\% & $\times$ 1.24 & 0.491 & 50\% & $\times$ 1.21 & -- & 50\% & -- \\
\multirow{2}{*}{\attnmarker\ PieceWise Attention}
& 0.455 & 25\% & $\times$ 0.73 & 0.483 & 25\% & $\times$ 0.68 & 0.486 & 25\% & $\times$ 0.69 \\
& 0.465 & 50\% & $\times$ 0.72 & 0.494 & 50\% & $\times$ 0.67 & 0.498 & 50\% & $\times$ 0.65 \\
\multirow{2}{*}{\mlpmarker\ ToMe}
& 0.403 & 25\% & $\times$ 0.83 & 0.425 & 25\% & $\times$ 0.86 & 0.430 & 25\% & $\times$ 0.80 \\
& 0.467 & 50\% & $\times$ 0.74 & 0.494 & 50\% & $\times$ 0.75 & 0.498 & 50\% & $\times$ 0.64 \\
\multirow{2}{*}{\mlpmarker\ StructSAM}
& 0.306 & 25\% & $\times$ 0.86 & 0.318 & 25\% & $\times$ 0.89 & 0.318 & 25\% & $\times$ 0.94 \\
& 0.422 & 50\% & $\times$ 0.71 & 0.445 & 50\% & $\times$ 0.74 & 0.451 & 50\% & $\times$ 0.79 \\
\hline
\multirow{2}{*}{\attnmarker\ SparseSAM (Ours)}
& \textbf{0.459} & 25\% & \textbf{$\times$ 1.63} & \textbf{0.487} & 25\% & \textbf{$\times$ 1.64} & \textbf{0.494} & 25\% & \textbf{$\times$ 1.28} \\
& \textbf{0.467} & 50\% & \textbf{$\times$ 1.55} & \textbf{0.494} & 50\% & \textbf{$\times$ 1.57} & \textbf{0.499} & 50\% & \textbf{$\times$ 1.22} \\
\multirow{2}{*}{\mlpmarker\ SparseSAM (Ours)}
& \textbf{0.451} & 25\% & \textbf{$\times$ 2.15} & \textbf{0.468} & 25\% & \textbf{$\times$ 2.05} & \textbf{0.472} & 25\% & \textbf{$\times$ 1.59} \\
& \textbf{0.459} & 50\% & \textbf{$\times$ 1.89} & \textbf{0.482} & 50\% & \textbf{$\times$ 1.83} & \textbf{0.487} & 50\% & \textbf{$\times$ 1.40} \\
\bottomrule
\end{tabular}}
\label{tab:coco-dino}
\end{table}
Table~\ref{tab:coco-dino} reports zero-shot segmentation results on MS-COCO~\cite{lin2014microsoft} using the DINO detector~\cite{zhang2022dino}. Across all settings, \name consistently achieves the best accuracy-efficiency trade-off, delivering the highest inference speed while preserving segmentation quality.

\textbf{Attention Comparison.}
SpargeAttention~\cite{zhang2025spargeattention} and PISA~\cite{li2026pisa} are limited by SAM's local attention design, where many attention heads operate on relatively small token maps (196 $\times$ 196). In this regime, the overhead of dynamic sparsification often outweighs the benefits of FlashAttention-style kernels~\cite{dao2023flashattention2}. PISA introduces additional Top-K selection and Taylor approximation overhead, making it slower than the dense baseline in some cases (e.g., 0.73$\times$ speedup on SAM-B at 25\% density). In contrast, \name\ reuses a fixed Z-curve permutation across all encoder layers, avoiding costly runtime operations. As a result, it achieves the highest efficiency, reaching 1.63$\times$ speedup on SAM-B at 25\% density, compared to 1.25$\times$ for SpargeAttention.

\textbf{Joint Attention and MLP Compression.}
\name\ also outperforms token-merging approaches such as ToMe~\cite{bolya2022token}, which suffer from inaccurate merge-and-unmerge operations that degrade segmentation fidelity. At 50\% density on SAM-L, ToMe achieves 0.425 mAP, while \name\ preserves 0.482 mAP and achieves up to 1.81$\times$ speedup.

\subsubsection{High Quality Segmentation}
\begin{figure}[t]
    \centering
    \includegraphics[width=\textwidth]{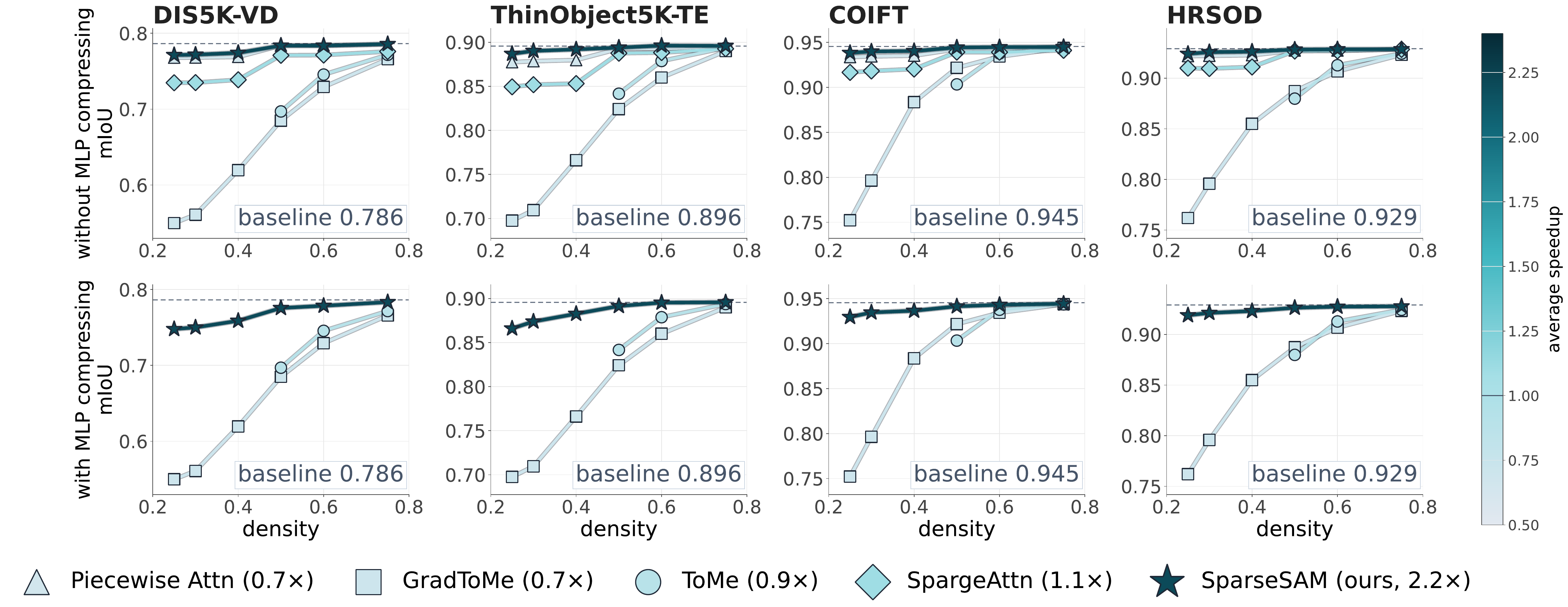}
    \caption{\textbf{Robustness Under Extreme Compression}. SparseSAM sets a new standard for efficient segmentation, consistently outperforming existing dynamic and merging-based approaches across density levels. Notably, at high compression rates (density $< 0.4$), where competing methods suffer catastrophic quality degradation, SparseSAM preserves near-baseline fidelity. This stability allows for a 2.2$\times$ end-to-end acceleration without the typical trade-off between inference speed and mask quality.}
    \label{fig:hq44k}
    \vspace{-.3cm}
\end{figure}

To better stress-test compression robustness, we evaluate \name on HQ-44K~\cite{ke2023segment}, which includes DIS5K-VD, ThinObject5K-TE, COIFT, and HRSOD, focusing on thin and small-object segmentation. As shown in Figure~\ref{fig:hq44k}, \name\ consistently preserves higher segmentation fidelity than both dynamic masking and token-reduction baselines.

\textbf{Attention-Only Analysis.} 
\textit{Under heavy compression}, our Stripe-Sort Attention remains highly stable. Leveraging deterministic Z-order interleaving and static A-shape masking, \name\ maintains performance close to the dense model even at 0.25 density (75\% sparsity), avoiding the collapse observed in dynamic masking methods. For example, on COIFT at 0.3 density, \name\ achieves 0.945 mIoU, outperforming SpargeAttention~\cite{zhang2025spargeattention} (0.919) and PISA~\cite{li2026pisa} (0.937). A similar gain is observed on ThinObject5K-TE, where \name\ reaches 0.892 mIoU, exceeding both PISA (0.878) and SpargeAttention (0.850).

\textbf{Joint Attention and MLP Compression.} 
With residual-consistent MLP compression, \name\ offers a strong accuracy–efficiency trade-off. At 0.5 density, it achieves 0.782 mIoU on DIS5K-VD, while ToMe~\cite{bolya2022token} (0.692) degrades substantially. Unlike token merging, our method preserves key information flow via residual propagation, allowing even low-rank tokens to contribute to later layers and preventing the information loss seen in merging-based approaches.

\subsection{Ablation Studies}

\paragraph{Finetune recovery.} Despite being a training-free method by design, we also attempt to recover the performance of the model by fine-tuning only the MLP layers of SparseSAM for 15 minutes under low-density settings. As shown in Table \ref{tab:finetune_recovery}, this simple adaptation consistently improves mIoU across datasets while keeping inference latency unchanged. At 25\% density, mIoU increases from 74.79 to 76.92 on DIS5K-VD and from 91.79 to 93.01 on COIFT, with similar gains at 50\% density. This demonstrates that minimal MLP fine-tuning effectively recovers accuracy under aggressive token reduction without sacrificing speed.
\begin{table}[h]
\centering
\definecolor{sparsegreen}{RGB}{28,155,138}
\newcommand{\mlpmarker}{\textcolor{sparsegreen}{\ensuremath{\bullet}}}
\caption{\textbf{Performance Recovery via Fine-tuning.} Fine-tuning SparseSAM at low densities (25\% and 50\%) significantly closes the mIoU gap while maintaining the same accelerated inference latency.}
\vspace{0.05in}
\small
\setlength{\tabcolsep}{5pt}
\resizebox{0.8\textwidth}{!}{
\begin{tabular}{l|c|cc|cc|cc}
\toprule
\multirow{2}{*}{Method} & \multirow{2}{*}{Density} & \multicolumn{2}{c|}{DIS5K-VD} & \multicolumn{2}{c|}{COIFT} & \multicolumn{2}{c}{HRSOD} \\
 & & mIoU & Time (ms) & mIoU & Time (ms) & mIoU & Time (ms) \\
\midrule
Base Model & 100\% & 78.63 & 55.02 & 94.55 & 50.23 & 92.91 & 54.67 \\
\midrule
\mlpmarker\ SparseSAM  & 25\% & 74.79 & 25.97 & 91.79 & 24.90 & 91.29 & 26.69 \\
\rowcolor{gray!15} \mlpmarker\ SparseSAM (+FT) & 25\% & \textbf{76.92} & 25.97 & \textbf{93.01} & 24.90 & \textbf{91.96} & 26.69 \\
\midrule
\mlpmarker\ SparseSAM (finetuned) & 50\% & 77.56 & 29.16 & 93.12 & 27.79 & 92.40 & 29.31 \\
\rowcolor{gray!15} \mlpmarker\ SparseSAM (+FT) & 50\% & \textbf{78.28} & 29.16 & \textbf{93.97} & 27.79 & \textbf{92.71} & 29.31 \\
\bottomrule
\end{tabular}}
\label{tab:finetune_recovery}
\vspace{-0.2in}
\end{table}

\begin{figure}[t]
    \centering
    \includegraphics[width=\textwidth]{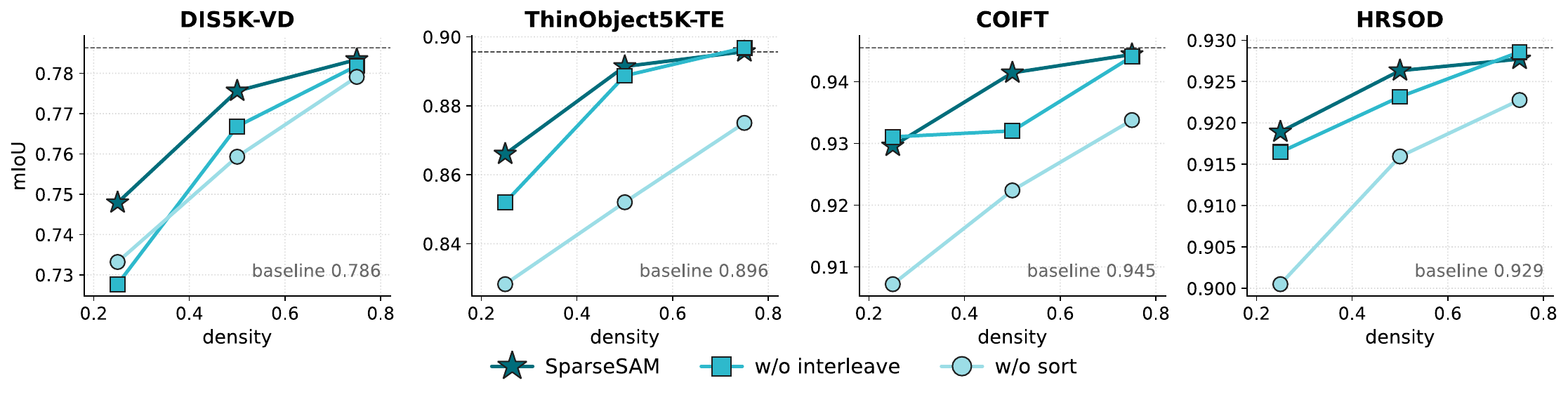}
    \caption{Performance breakdown with different permutation orders.}
    \label{fig:perm_ablation}
    \vspace{-0.25in}
\end{figure}

\paragraph{Different permutation orders in Sparse Attention kernel.}
Our token permutation combines (i) \emph{Gradient-Based Ordering}, which prioritizes high-importance Z-curve groups, and (ii) \emph{Stripe-Sort Permutation}, which interleaves groups to maintain spatially uniform kept tokens. We evaluate two ablations on SAM-HQ ViT-L over HQ44K: \emph{w/o interleave}, which removes interleaving while preserving ranking, and \emph{w/o sort}, which preserves interleaving but removes content-aware ranking. As shown in Figure~\ref{fig:perm_ablation}, all variants perform similarly at mild compression ($r=0.75$), but differences become significant at lower densities. At $r=0.25$, removing interleaving reduces performance by up to $-0.020$ mIoU on DIS5K-VD, particularly for thin structures. Removing ranking causes larger drops, especially on ThinObject5K ($-0.038$) and COIFT ($-0.022$), indicating that content-aware prioritization is the primary factor under aggressive compression.

\begin{wraptable}{r}{0.6\textwidth}
    \vspace{-13pt}
    \centering
    \caption{SparseSAM applied to SAM2 with the HQ-Hiera-L encoder.}
    \label{tab:sam2-hq44k}
    \setlength{\tabcolsep}{4pt}
    \footnotesize
    \begin{tabular}{@{}c cc cc cc cc@{}}
        \toprule
        & \multicolumn{2}{c}{COIFT} & \multicolumn{2}{c}{DIS5K-VD} & \multicolumn{2}{c}{HRSOD} & \multicolumn{2}{c}{ThinObject5K} \\
        \cmidrule(lr){2-3}\cmidrule(lr){4-5}\cmidrule(lr){6-7}\cmidrule(lr){8-9}
        $r$ & mIoU & bIoU & mIoU & bIoU & mIoU & bIoU & mIoU & bIoU \\
        \midrule
        1.00 & 0.951 & 0.908 & 0.753 & 0.670 & 0.909 & 0.854 & 0.903 & 0.819 \\
        0.70 & 0.951 & 0.907 & 0.754 & 0.674 & 0.912 & 0.858 & 0.906 & 0.825 \\
        0.50 & 0.951 & 0.906 & 0.761 & 0.677 & 0.916 & 0.861 & 0.913 & 0.834 \\
        0.30 & 0.946 & 0.901 & 0.767 & 0.678 & 0.925 & 0.867 & 0.912 & 0.824 \\
        \bottomrule
    \end{tabular}
\end{wraptable}

\paragraph{Can residual-consistency MLP be used on other segmentation backbones?}
To further confirm the generality of our residual-consistency MLP, we apply SparseSAM to SAM2, where the tokens passing through each MLP layer are routed by the same residual-consistency mechanism. As shown in Table~\ref{tab:sam2-hq44k}, our observation transfers cleanly to the Hiera encoder of SAM2, whose backbone likewise exhibits distinct token semantics across different regions of the image. We also attempted to adapt SparseSAM to other backbone that was train on contrastive learning objective and evaluate on Imagenet \cite{deng2009imagenet}  for more information, please refer to \ref{sec:MLP_behavior} and \ref{sec:imagenet_res}.    

\vspace{-0.05in}
\section{Conclusion}
\vspace{-0.05in}
We introduced SparseSAM, a training-free framework that accelerates the Segment Anything Model and its successors. By combining Stripe-Sort Permutation to exploit spatial redundancy with a Residual-Consistency MLP to preserve critical information flow, SparseSAM achieves a 2$\times$ inference speedup and 2.8$\times$ memory reduction with negligible fidelity loss. Validated across five diverse datasets, our approach maintains high performance while enabling the deployment of large-scale segmentation foundation models on resource-constrained edge devices.

For future work, several promising avenues remain. First, since our deterministic Z-order permutation is content-agnostic, exploring hybrid strategies could allow the model to dynamically prioritize semantic importance, better capturing long-range dependencies in complex scenes without losing the efficiency of static sparsity. Second, we aim to investigate layer-adaptive routing for the Residual-Consistency MLP to better account for evolving feature hierarchies as tokens transition from low-level textures to high-level representations. Finally, porting SparseSAM to backends like OpenCL or specialized DSP instructions will facilitate high-performance deployment across a wider array of mobile and edge devices.

\begin{ack}
This work was supported by the Siebel School of Computing and Data Science at the University of Illinois Urbana-Champaign (UIUC) and by the VinUni-Illinois Smart Health Center (VISHC), VinUniversity. The authors also thank the International Max Planck Research School for Intelligent Systems (IMPRS-IS) for providing feedback and additional computing resources.
\end{ack}

\bibliographystyle{plain}
\bibliography{ref}
\newpage

\appendix

\section{Appendix}

\subsection{Attention Kernel Pseudo Code}
\label{sec:flash_kernel}
\begin{algorithm}[H]
\caption{A-Shape Windowed FlashAttention-2 with Decomposed 2D Rel-Pos Bias}
\begin{algorithmic}
\algcomment{Attention input with decomposed 2D bias $\mathbf{B}[q,k]\!=\!\mathbf{B}^{\!H}[q,k_{\text{row}}]\!+\!\mathbf{B}^{\!W}[q,k_{\text{col}}]$ on a $w\!\times\!w$ grid}
\State \textbf{Input:}\, $\mathbf{Q}\!\in\!\mathbb{R}^{S_q\times d}$,\, $\mathbf{K},\mathbf{V}\!\in\!\mathbb{R}^{S_k\times d}$;\;\, $\mathbf{B}^{\!H},\mathbf{B}^{\!W}\!\in\!\mathbb{R}^{S_q\times w}$,\, $w\!=\!\sqrt{S_k}$
\algcomment{Window-major $\to$ spatial perms (for bias lookup); A-mask density ratio; softmax scale}
\State $\sigma_Q\!:\![S_q]\!\to\![S_q]$,\, $\sigma_K\!:\![S_k]\!\to\![S_k]$;\;\, $r\!\in\![0,1]$;\;\, $\tau\!=\!1/\!\sqrt{d}$
\algcomment{Tile sizes, counts, slices, tile-local indices}
\State \textbf{Tiling:}\, $B_{\text{row}},B_{\text{col}}$;\, $T_{\text{row}}\!=\!\lceil S_q/B_{\text{row}}\rceil$,\, $T_{\text{col}}\!=\!\lceil S_k/B_{\text{col}}\rceil$;\, $\mathbf{Q}_i\!\triangleq\!\mathbf{Q}[iB_{\text{row}}{:}(i{+}1)B_{\text{row}}]$,\, $\mathbf{K}_j,\mathbf{V}_j$;\, $\text{row}\!\in\![0,B_{\text{row}})$,\, $\text{col}\!\in\![0,B_{\text{col}})$
\State \textbf{Output:}\, $\mathbf{O}\!\in\!\R^{S_q\times d}$
\algcomment{Active K-tiles: first $\lfloor r T_{\text{col}}\rfloor$ columns plus the diagonal $j\!=\!i$}
\State $\mathcal{J}_i \leftarrow \{0,\dots,\lfloor r T_{\text{col}}\rfloor\!-
1\}\cup\{i\}$
\algcomment{Gather bias rows: query at tile-local row $\text{row}$ has spatial position $\sigma_Q(iB_{\text{row}}{+}\text{row})$}
\State $\mathbf{B}^{\!H}_{\!i}[\text{row},p] \leftarrow \mathbf{B}^{\!H}[\sigma_Q(iB_{\text{row}}{+}\text{row}),p]$;\;\; $\mathbf{B}^{\!W}_{\!i}[\text{row},p] \leftarrow \mathbf{B}^{\!W}[\sigma_Q(iB_{\text{row}}{+}\text{row}),p]$
\algcomment{Online-softmax state: per-row max, denom, output accumulator}
\State $\mathbf{m} \leftarrow -\boldsymbol{\infty}$,\;\; $\boldsymbol{\ell} \leftarrow \mathbf{0}$,\;\; $\mathbf{O}_{\!\text{acc}} \leftarrow \mathbf{0}$
\For{$j \in \mathcal{J}_i$}
    \State $\mathbf{S} \leftarrow \mathbf{Q}_i\,\mathbf{K}_j^{\!\top}$
    \algcomment{Add bias; $\div\tau$ cancels the $\tau$ in the softmax exponent below}
    \State $\mathbf{S}[\text{row},\text{col}] \mathrel{+}= \big(\mathbf{B}^{\!H}_{\!i}[\text{row},\,\sigma_K(\text{col})\!\div\! w] + \mathbf{B}^{\!W}_{\!i}[\text{row},\,\sigma_K(\text{col})\!\bmod\! w]\big)/\tau$
    \State $\mathbf{S}[\text{row},\text{col}] \leftarrow -\infty$ \textbf{where} $jB_{\text{col}}+\text{col}\ge S_k$
    \algcomment{FA online-softmax update}
    \State $\mathbf{m}' \leftarrow \max(\mathbf{m},\operatorname{rowmax}\mathbf{S})$;\;\; $\widetilde{\mathbf{P}} \leftarrow \exp(\tau(\mathbf{S}-\mathbf{m}'))$;\;\; $\boldsymbol{\alpha} \leftarrow \exp(\tau(\mathbf{m}-\mathbf{m}'))$
    \State $\boldsymbol{\ell} \leftarrow \boldsymbol{\alpha}\!\odot\!\boldsymbol{\ell} + \operatorname{rowsum}\widetilde{\mathbf{P}}$;\;\; $\mathbf{O}_{\!\text{acc}} \leftarrow \operatorname{diag}(\boldsymbol{\alpha})\mathbf{O}_{\!\text{acc}} + \widetilde{\mathbf{P}}\mathbf{V}_j$;\;\; $\mathbf{m}\leftarrow\mathbf{m}'$
\EndFor
\State $\mathbf{O}_i \leftarrow \operatorname{diag}(\boldsymbol{\ell})^{-1}\mathbf{O}_{\!\text{acc}}$
\end{algorithmic}
\end{algorithm}

In SAM-based models, the attention operator includes a non-trivial relative positional bias that is directly added to the attention logits. To efficiently integrate a customized sparse attention kernel, we fuse this operation into a FlashAttention-2-style computation with decomposed 2D positional encoding, where the bias is factorized as
$\mathbf{B}[q,k] = \mathbf{B}^{H}[q,k_{\text{row}}] + \mathbf{B}^{W}[q,k_{\text{col}}]$ on a $w \times w$ grid with $w=\sqrt{S_k}$.

The input features $\mathbf{Q}\in\mathbb{R}^{S_q\times d}$, $\mathbf{K},\mathbf{V}\in\mathbb{R}^{S_k\times d}$ are first reordered using spatial permutations $\sigma_Q$ and $\sigma_K$, which improve memory locality under a window-major tiling scheme. Attention is computed in a block-wise manner over tiles $\mathbf{Q}_i$ and $\mathbf{K}_j$, with tile sizes $B_{\text{row}}$ and $B_{\text{col}}$, and only a subset of key blocks $\mathcal{J}_i$ is evaluated according to a structured A-shaped sparsity mask (controlled by density ratio $r$), which is statically compiled and incurs no runtime overhead.

For each tile interaction, attention scores are computed as $\mathbf{S}=\mathbf{Q}_i\mathbf{K}_j^\top$, and the decomposed positional bias is added via efficient index-based lookup using $\mathbf{B}^{H}$ and $\mathbf{B}^{W}$ after applying $\sigma_Q$ and $\sigma_K$. Entries outside valid key boundaries are masked to $-\infty$.

The kernel follows the FlashAttention-2 online softmax formulation, maintaining running statistics $\mathbf{m}$ (row-wise maxima), $\boldsymbol{\ell}$ (normalization terms), and $\mathbf{O}_{\text{acc}}$ (output accumulator). These are updated incrementally across key tiles to ensure numerical stability while streaming. The final output $\mathbf{O}_i$ is obtained by normalizing $\mathbf{O}_{\text{acc}}$ with $\boldsymbol{\ell}$ after processing all active blocks in $\mathcal{J}_i$.

\subsection{Full MS-COCO Results}
\begin{table}[t]
\centering
\definecolor{sparsegreen}{RGB}{28,155,138}
\definecolor{finetunegray}{gray}{0.92}
\newcommand{\attnmarker}{\textcolor{black}{\ensuremath{\circ}}}
\newcommand{\mlpmarker}{\textcolor{sparsegreen}{\ensuremath{\bullet}}}
\caption{MS-COCO results. Markers indicate compression type: \protect\attnmarker\ attention-only and \protect\mlpmarker\ attention+MLP.}
\scriptsize
\setlength{\tabcolsep}{2pt}
\renewcommand{\arraystretch}{1.05}
\resizebox{0.85\textwidth}{!}{
\begin{tabular}{l|l|ccc|ccc|ccc}
\toprule
\multirow{2}{*}{Detector} & \multirow{2}{*}{Method} & \multicolumn{3}{c|}{SAM-B} & \multicolumn{3}{c|}{SAM-L} & \multicolumn{3}{c}{SAM-H} \\
& & mAP & Density & Speedup & mAP & Density & Speedup & mAP & Density & Speedup \\
\midrule
\multirow{16}{*}{DINO}
& Base & 0.468 & 100\% & $\times$1.00 & 0.495 & 100\% & $\times$1.00 & 0.499 & 100\% & $\times$1.00 \\
\cmidrule(lr){2-11}
& \multicolumn{10}{l}{\textit{Attention-only compression}} \\
\cmidrule(lr){2-11}
& \multirow{2}{*}{\attnmarker\ Sparge Attn} & 0.447 & 25\% & $\times$1.25 & 0.461 & 25\% & $\times$1.23 & -- & 25\% & -- \\
& & 0.463 & 50\% & $\times$1.24 & 0.491 & 50\% & $\times$1.21 & -- & 50\% & -- \\
& \multirow{2}{*}{\attnmarker\ PieceWise Attn} & 0.455 & 25\% & $\times$0.73 & 0.483 & 25\% & $\times$0.68 & 0.486 & 25\% & $\times$0.69 \\
& & 0.465 & 50\% & $\times$0.72 & 0.494 & 50\% & $\times$0.67 & 0.498 & 50\% & $\times$0.65 \\
\cmidrule(lr){2-11}
& \multirow{2}{*}{\attnmarker\ SparseSAM (Ours)} & \textbf{0.459} & 25\% & \textbf{$\times$1.63} & \textbf{0.487} & 25\% & \textbf{$\times$1.64} & \textbf{0.494} & 25\% & \textbf{$\times$1.28} \\
& & \textbf{0.467} & 50\% & \textbf{$\times$1.55} & \textbf{0.494} & 50\% & \textbf{$\times$1.57} & \textbf{0.499} & 50\% & \textbf{$\times$1.22} \\
\cmidrule(lr){2-11}
& \multicolumn{10}{l}{\textit{Attention + MLP compression}} \\
\cmidrule(lr){2-11}
& \mlpmarker\ ToMe & 0.403 & 50\% & $\times$0.83 & 0.425 & 50\% & $\times$0.86 & 0.430 & 50\% & $\times$0.80 \\
& \multirow{2}{*}{\mlpmarker\ GradToMe} & 0.306 & 25\% & $\times$0.86 & 0.318 & 25\% & $\times$0.89 & 0.318 & 25\% & $\times$0.94 \\
& & 0.422 & 50\% & $\times$0.71 & 0.445 & 50\% & $\times$0.74 & 0.451 & 50\% & $\times$0.79 \\
\cmidrule(lr){2-11}
& \multirow{2}{*}{\mlpmarker\ SparseSAM (Ours)} & \textbf{0.451} & 25\% & \textbf{$\times$2.15} & \textbf{0.468} & 25\% & \textbf{$\times$2.05} & \textbf{0.472} & 25\% & \textbf{$\times$1.59} \\
& & \textbf{0.459} & 50\% & \textbf{$\times$1.89} & \textbf{0.482} & 50\% & \textbf{$\times$1.83} & \textbf{0.487} & 50\% & \textbf{$\times$1.40} \\
\midrule
\multirow{16}{*}{HDETR}
& Base & 0.402 & 100\% & $\times$1.00 & 0.424 & 100\% & $\times$1.00 & 0.427 & 100\% & $\times$1.00 \\
\cmidrule(lr){2-11}
& \multicolumn{10}{l}{\textit{Attention-only compression}} \\
\cmidrule(lr){2-11}
& \multirow{2}{*}{\attnmarker\ Sparge Attn} & 0.386 & 25\% & $\times$1.25 & 0.398 & 25\% & $\times$1.20 & -- & 25\% & -- \\
& & 0.398 & 50\% & $\times$1.24 & 0.421 & 50\% & $\times$1.18 & -- & 50\% & -- \\
& \multirow{2}{*}{\attnmarker\ PieceWise Attn} & 0.391 & 25\% & $\times$0.72 & 0.415 & 25\% & $\times$0.65 & 0.418 & 25\% & $\times$0.68 \\
& & 0.401 & 50\% & $\times$0.70 & 0.422 & 50\% & $\times$0.63 & 0.426 & 50\% & $\times$0.64 \\
\cmidrule(lr){2-11}
& \multirow{2}{*}{\attnmarker\ SparseSAM (Ours)} & \textbf{0.396} & 25\% & \textbf{$\times$1.73} & \textbf{0.418} & 25\% & \textbf{$\times$1.63} & \textbf{0.423} & 25\% & \textbf{$\times$1.27} \\
& & \textbf{0.401} & 50\% & \textbf{$\times$1.66} & \textbf{0.423} & 50\% & \textbf{$\times$1.56} & \textbf{0.427} & 50\% & \textbf{$\times$1.20} \\
\cmidrule(lr){2-11}
& \multicolumn{10}{l}{\textit{Attention + MLP compression}} \\
\cmidrule(lr){2-11}
& \mlpmarker\ ToMe & 0.351 & 50\% & $\times$0.83 & 0.369 & 50\% & $\times$0.86 & 0.373 & 50\% & $\times$0.78 \\
& \multirow{2}{*}{\mlpmarker\ GradToMe} & 0.271 & 25\% & $\times$0.86 & 0.280 & 25\% & $\times$0.88 & 0.280 & 25\% & $\times$0.93 \\
& & 0.366 & 50\% & $\times$0.70 & 0.386 & 50\% & $\times$0.72 & 0.390 & 50\% & $\times$0.78 \\
\cmidrule(lr){2-11}
& \multirow{2}{*}{\mlpmarker\ SparseSAM (Ours)} & \textbf{0.389} & 25\% & \textbf{$\times$2.16} & \textbf{0.403} & 25\% & \textbf{$\times$2.05} & \textbf{0.406} & 25\% & \textbf{$\times$1.58} \\
& & \textbf{0.394} & 50\% & \textbf{$\times$1.91} & \textbf{0.415} & 50\% & \textbf{$\times$1.83} & \textbf{0.417} & 50\% & \textbf{$\times$1.37} \\
\midrule
\multirow{16}{*}{YOLOX}
& Base & 0.390 & 100\% & $\times$1.00 & 0.412 & 100\% & $\times$1.00 & 0.416 & 100\% & $\times$1.00 \\
\cmidrule(lr){2-11}
& \multicolumn{10}{l}{\textit{Attention-only compression}} \\
\cmidrule(lr){2-11}
& \multirow{2}{*}{\attnmarker\ Sparge Attn} & 0.374 & 25\% & $\times$1.25 & 0.386 & 25\% & $\times$1.20 & -- & 25\% & -- \\
& & 0.386 & 50\% & $\times$1.24 & 0.409 & 50\% & $\times$1.18 & -- & 50\% & -- \\
& \multirow{2}{*}{\attnmarker\ PieceWise Attn} & 0.379 & 25\% & $\times$0.71 & 0.403 & 25\% & $\times$0.65 & 0.406 & 25\% & $\times$0.67 \\
& & 0.388 & 50\% & $\times$0.69 & 0.410 & 50\% & $\times$0.63 & 0.414 & 50\% & $\times$0.63 \\
\cmidrule(lr){2-11}
& \multirow{2}{*}{\attnmarker\ SparseSAM (Ours)} & \textbf{0.384} & 25\% & \textbf{$\times$1.70} & \textbf{0.406} & 25\% & \textbf{$\times$1.63} & \textbf{0.411} & 25\% & \textbf{$\times$1.29} \\
& & \textbf{0.390} & 50\% & \textbf{$\times$1.63} & \textbf{0.410} & 50\% & \textbf{$\times$1.56} & \textbf{0.415} & 50\% & \textbf{$\times$1.23} \\
\cmidrule(lr){2-11}
& \multicolumn{10}{l}{\textit{Attention + MLP compression}} \\
\cmidrule(lr){2-11}
& \mlpmarker\ ToMe & 0.343 & 50\% & $\times$0.83 & 0.361 & 50\% & $\times$0.86 & 0.365 & 50\% & $\times$0.79 \\
& \multirow{2}{*}{\mlpmarker\ GradToMe} & 0.267 & 25\% & $\times$0.85 & 0.276 & 25\% & $\times$0.88 & 0.276 & 25\% & $\times$0.92 \\
& & 0.356 & 50\% & $\times$0.69 & 0.377 & 50\% & $\times$0.72 & 0.381 & 50\% & $\times$0.77 \\
\cmidrule(lr){2-11}
& \multirow{2}{*}{\mlpmarker\ SparseSAM (Ours)} & \textbf{0.377} & 25\% & \textbf{$\times$2.15} & \textbf{0.391} & 25\% & \textbf{$\times$2.05} & \textbf{0.395} & 25\% & \textbf{$\times$1.58} \\
& & \textbf{0.384} & 50\% & \textbf{$\times$1.92} & \textbf{0.403} & 50\% & \textbf{$\times$1.83} & \textbf{0.405} & 50\% & \textbf{$\times$1.40} \\
\bottomrule
\end{tabular}}
\label{tab:coco}
\end{table}
Table~\ref{tab:coco} presents a detailed performance comparison of SparseSAM across the MS-COCO~\cite{lin2014microsoft} dataset, utilizing SAM-B, SAM-L, and SAM-H backbones under three distinct detectors: DINO~\cite{zhang2022dino}, H-DETR~\cite{jia2022detrs}, and YOLOX~\cite{ge2021yolox}. The results indicate that SparseSAM consistently outperforms existing attention-only and token-reduction baselines in both segmentation accuracy (mAP) and inference efficiency. Notably, under joint attention and MLP compression, our method achieves state-of-the-art speedups—reaching up to 2.16$\times$ at 25\% density—while maintaining mAP scores remarkably close to the dense base models. This demonstrates the robustness of SparseSAM's structured sparsity across various model scales and detection frameworks, significantly mitigating the drastic accuracy drops observed in merging-based approaches like ToMe~\cite{bolya2022token}.
\subsection{Full HQ44k Results}
\begin{table}[t]
\centering
\definecolor{sparsegreen}{RGB}{28,155,138}
\definecolor{sparseblue}{RGB}{0,114,189}
\newcommand{\attnmarker}{\textcolor{black}{\ensuremath{\circ}}}
\newcommand{\mlpmarker}{\textcolor{sparsegreen}{\ensuremath{\bullet}}}
\newcommand{\finemarker}{\textcolor{sparseblue}{\ensuremath{\bullet}}}
\caption{Segmentation results. Markers indicate compression type: \protect\attnmarker\ attention-only, \protect\mlpmarker\ attention+MLP, and \protect\finemarker\ denotes finetuned models.}
\scriptsize
\setlength{\tabcolsep}{4pt}
\renewcommand{\arraystretch}{1.12}
\resizebox{\textwidth}{!}{%
\begin{tabular}{l|c|cc|cc|cc|cc}
\toprule
\multirow{2}{*}{Method} & \multirow{2}{*}{Density} & \multicolumn{2}{c|}{DIS5K-VD} & \multicolumn{2}{c|}{COIFT} & \multicolumn{2}{c|}{ThinObject5K-TE} & \multicolumn{2}{c}{HRSOD} \\
 & & mIoU & Speedup & mIoU & Speedup & mIoU & Speedup & mIoU & Speedup \\
\midrule
Base Model & 100\% & 78.63 & $\times$1.00 & 94.55 & $\times$1.00 & 89.56 & $\times$1.00 & 92.91 & $\times$1.00 \\
\midrule
\multicolumn{10}{c}{\textit{Attention-only Sparsify}} \\
\midrule
\multirow{3}{*}{\attnmarker\ SpargeAttn}
 & 25\% & 73.48 & $\times$1.26 & 91.67 & $\times$1.25 & 84.94 & $\times$1.24 & 90.99 & $\times$1.25 \\
 & 50\% & 77.09 & $\times$1.24 & 93.92 & $\times$1.23 & 88.76 & $\times$1.25 & 92.68 & $\times$1.24 \\
 & 75\% & 77.61 & $\times$1.16 & 94.12 & $\times$1.14 & 89.26 & $\times$1.15 & 92.84 & $\times$1.15 \\
\addlinespace[1pt]
\multirow{3}{*}{\attnmarker\ PieceWise Attention}
 & 25\% & 76.79 & $\times$0.71 & 93.37 & $\times$0.70 & 87.74 & $\times$0.72 & 92.18 & $\times$0.70 \\
 & 50\% & 78.38 & $\times$0.69 & 94.25 & $\times$0.71 & 89.26 & $\times$0.70 & 92.73 & $\times$0.71 \\
 & 75\% & 78.54 & $\times$0.68 & 94.51 & $\times$0.67 & 89.49 & $\times$0.67 & 92.87 & $\times$0.66 \\
\hline
\multirow{3}{*}{\attnmarker\ SparseSAM (Ours)}
 & 25\% & \textbf{77.13} & \textbf{$\times$1.75} & \textbf{93.83} & \textbf{$\times$1.63} & \textbf{88.68} & \textbf{$\times$1.68} & \textbf{92.41} & \textbf{$\times$1.71} \\
 & 50\% & \textbf{78.38} & \textbf{$\times$1.67} & \textbf{94.44} & \textbf{$\times$1.56} & \textbf{89.40} & \textbf{$\times$1.61} & \textbf{92.83} & \textbf{$\times$1.63} \\
 & 75\% & \textbf{78.58} & \textbf{$\times$1.61} & \textbf{94.49} & \textbf{$\times$1.50} & \textbf{89.58} & \textbf{$\times$1.55} & \textbf{92.86} & \textbf{$\times$1.58} \\
\midrule
\multicolumn{10}{c}{\textit{Attention + MLP compression}} \\
\midrule
\multirow{2}{*}{\mlpmarker\ ToMe}
 & 50\% & 69.69 & $\times$0.81 & 90.33 & $\times$0.81 & 84.17 & $\times$0.81 & 87.98 & $\times$0.84 \\
 & 75\% & 77.16 & $\times$0.52 & 94.36 & $\times$0.51 & 89.38 & $\times$0.52 & 92.43 & $\times$0.51 \\
\addlinespace[1pt]
\multirow{3}{*}{\mlpmarker\ GradToMe}
 & 25\% & 54.97 & $\times$0.74 & 75.22 & $\times$0.75 & 69.76 & $\times$0.75 & 76.19 & $\times$0.77 \\
 & 50\% & 68.50 & $\times$0.63 & 92.17 & $\times$0.64 & 82.42 & $\times$0.64 & 88.73 & $\times$0.63 \\
 & 75\% & 76.58 & $\times$0.53 & 94.37 & $\times$0.54 & 89.00 & $\times$0.55 & 92.32 & $\times$0.54 \\
\hline
\multirow{3}{*}{\mlpmarker\ SparseSAM (Ours)}
 & 25\% & \textbf{74.79} & \textbf{$\times$2.12} & \textbf{91.79} & \textbf{$\times$2.02} & \textbf{85.95} & \textbf{$\times$2.08} & \textbf{91.29} & \textbf{$\times$2.05} \\
 & 50\% & \textbf{77.56} & \textbf{$\times$1.89} & \textbf{93.12} & \textbf{$\times$1.81} & \textbf{87.82} & \textbf{$\times$1.88} & \textbf{92.40} & \textbf{$\times$1.87} \\
 & 75\% & \textbf{78.43} & \textbf{$\times$1.73} & \textbf{94.12} & \textbf{$\times$1.64} & \textbf{89.21} & \textbf{$\times$1.70} & \textbf{92.62} & \textbf{$\times$1.71} \\
\midrule
\multicolumn{10}{c}{\textit{Finetuning after Joint Compression}} \\
\midrule
\finemarker\ SparseSAM (Ours) & 25\% & \textbf{76.92} & \textbf{$\times$2.12} & \textbf{93.01} & \textbf{$\times$2.02} & \textbf{88.05} & \textbf{$\times$2.08} & \textbf{91.96} & \textbf{$\times$2.05} \\
\finemarker\ SparseSAM (Ours) & 50\% & \textbf{78.28} & \textbf{$\times$1.89} & \textbf{93.97} & \textbf{$\times$1.81} & \textbf{89.41} & \textbf{$\times$1.88} & \textbf{92.71} & \textbf{$\times$1.87} \\
\bottomrule
\end{tabular}%
}
\label{tab:hq44k}
\end{table}

Table~\ref{tab:hq44k} presents an extensive evaluation of SparseSAM’s performance on zero-shot object segmentation across four challenging datasets including DIS5K-VD, COIFT, ThinObject5K-TE, and HRSOD from HQ-44K~\cite{ke2023segment}. The results demonstrate that SparseSAM maintains high segmentation accuracy while achieving superior speedup compared to existing training-free compression methods like SpargeAttention~\cite{zhang2025spargeattention} and PISA~\cite{li2026pisa}. Under joint attention and MLP compression at 25\% density, the model reaches peak speedup factors exceeding 2.0$\times$ while suffering significantly less accuracy degradation than competitive methods such as ToMe~\cite{bolya2022token}. The performance on ThinObject5K-TE particularly underscores the method's ability to preserve intricate spatial details even at low density levels. 

The final tier of the evaluation (blue markers) demonstrates that a lightweight finetuning stage successfully bridges the performance gap inherent in high-compression regimes. By allowing the model to adapt to the structured sparsity patterns in both the attention and MLP layers, SparseSAM recovers mIoU to near-base levels while fully preserving its throughput advantages. Specifically, the finetuned configurations at 25\% and 50\% density show negligible performance degradation compared to the 100\% density baseline. This underscores the efficacy of our approach in delivering a robust, hardware-efficient solution that maintains high-fidelity segmentation without sacrificing state-of-the-art inference acceleration.

\subsection{Hardware-Aware Performance Analysis}
\begin{table*}[t]
\centering
\caption{Throughput speedup comparison on different hardware and sparsity levels. We evaluate SparseSAM under two configurations: attention-only compression and joint attention + MLP compression.}
\label{tab:throughput-speedup}
\small
\resizebox{\textwidth}{!}{%
\begin{tabular}{ll|cc|cc|cc}
\toprule
\multirow{2}{*}{Batchsize} & \multirow{2}{*}{GPU} & \multicolumn{2}{c|}{25\% Sparsity} & \multicolumn{2}{c|}{50\% Sparsity} & \multicolumn{2}{c}{75\% Sparsity} \\
& & Attn. + MLP & Attn. Only & Attn. + MLP & Attn. Only & Attn. + MLP & Attn. Only \\
\midrule
\multirow{2}{*}{1} & A100 & $\times$1.91 & $\times$1.65 & $\times$1.75 & $\times$1.56 & $\times$1.33 & $\times$1.23 \\
                   & RTX 3090 & $\times$1.42 & $\times$1.32 & $\times$1.35 & $\times$1.25 & $\times$1.17 & $\times$1.15 \\
\midrule
\multirow{2}{*}{2} & A100 & $\times$1.95 & $\times$1.62 & $\times$1.75 & $\times$1.55 & $\times$1.33 & $\times$1.22 \\
                   & RTX 3090 & $\times$1.45 & $\times$1.33 & $\times$1.36 & $\times$1.25 & $\times$1.16 & $\times$1.16 \\
\midrule
\multirow{2}{*}{4} & A100 & $\times$2.00 & $\times$1.62 & $\times$1.80 & $\times$1.55 & $\times$1.46 & $\times$1.42 \\
                   & RTX 3090 & $\times$1.53 & $\times$1.33 & $\times$1.39 & $\times$1.26 & $\times$1.23 & $\times$1.27 \\
\midrule
\multirow{2}{*}{8} & A100 & $\times$2.05 & $\times$1.64 & $\times$1.83 & $\times$1.57 & $\times$1.66 & $\times$1.50 \\
                   & RTX 3090 & $\times$1.57 & $\times$1.34 & $\times$1.45 & $\times$1.27 & $\times$1.28 & $\times$1.28 \\
\bottomrule
\end{tabular}%
}
\label{tab:speed}
\end{table*}
Table~\ref{tab:speed} provides a detailed throughput speedup analysis of SparseSAM, comparing the performance of attention-only compression against joint attention and MLP compression across A100 and RTX 3090 GPUs. The results reveal that incorporating our Residual-Preserving MLP technique consistently enhances efficiency, yielding superior throughput compared to attention-only masking across all tested density ratios. Notably, the A100 architecture achieves a peak speedup of 2.05$\times$ at 25\% sparsity, demonstrating higher hardware utilization than the RTX 3090. Furthermore, the speedup gains for the joint compression configuration scale positively with larger batch sizes, validating that our structured sparsity approach effectively maximizes computational throughput while maintaining the benefits of training-free acceleration.

\subsection{MLP update behavior in other global task backbones} \label{sec:MLP_behavior}
\begin{figure}[t]
    \centering
    \includegraphics[width=\textwidth]{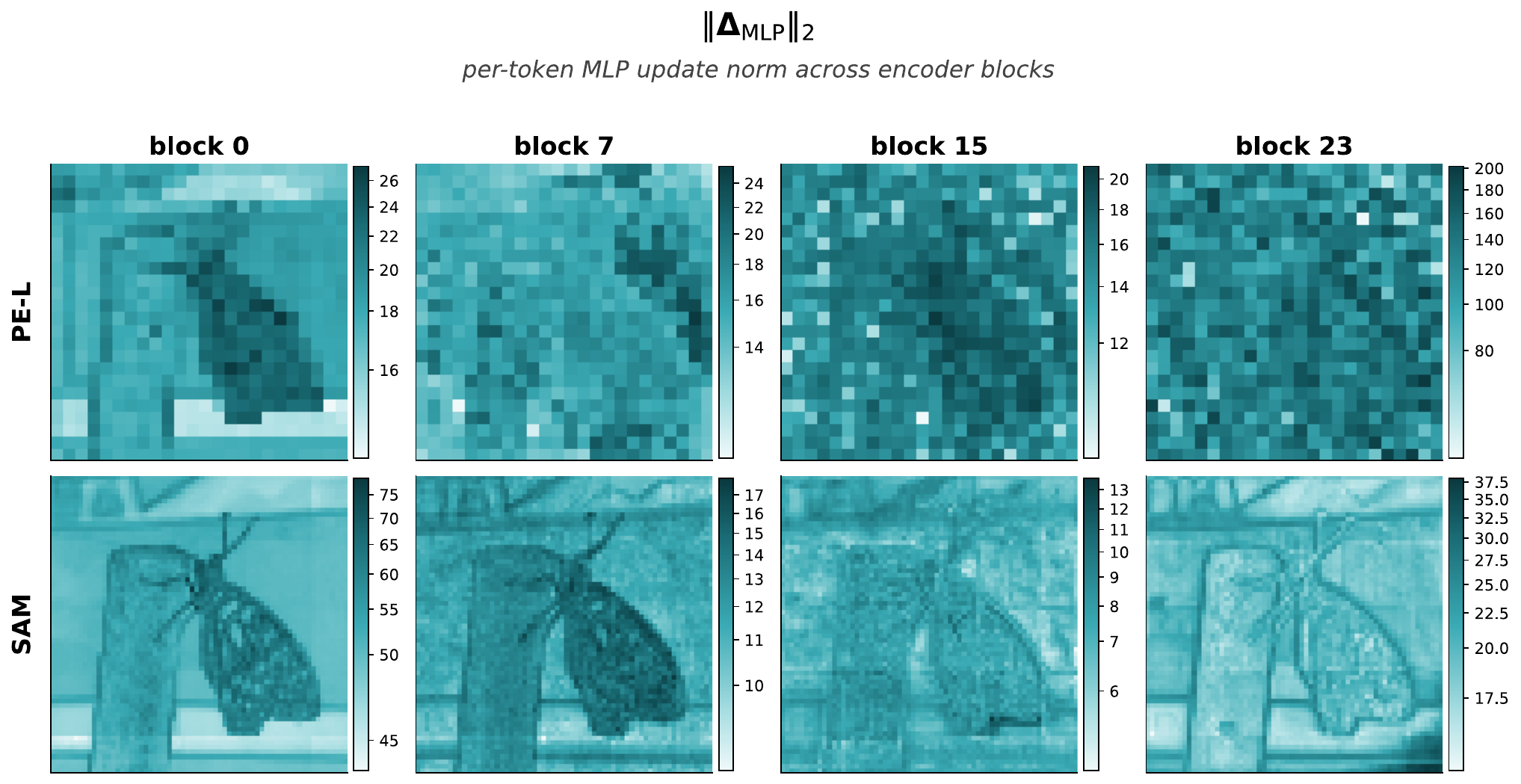}
    \caption{ Per-token MLP update norm $||\boldsymbol{\Delta}_{\mathrm{MLP}}||_2$ at blocks 0/7/15/23 of PE-Core-L14-336 (top) and SAM-HQ ViT-L (bottom) on the same input. Reflecting their different pretraining inductive biases, the two backbones distribute MLP work in opposite ways. The first block looks similar across models — both focus updates on the butterfly — but in later layers, PE updates tokens uniformly across the image, while SAM stays spatially selective and keeps its MLP work concentrated on the foreground.}
    \label{fig:mlp_update}
\end{figure}

To understand how MLPs distribute work across an encoder, we visualize the per-token MLP update norm $\|\boldsymbol{\Delta}_{\mathrm{MLP}}\|_2 = \|\mathrm{MLP}(\mathrm{LN}(x))\|_2$ at four representative blocks of Perception Encoder \cite{bolya2025perception} (PE-Core-L14-336) and SAM-L on the same image. From Figure \ref{fig:mlp_update}, it can be seen that the two backbones use their MLPs in markedly different ways. At block 0, behavior is broadly aligned: both models drive larger updates on the butterfly than on the background, indicating early object-aware encoding. Beyond the first block, the patterns diverge. PE, trained with an image-level contrastive objective, gradually loses spatial selectivity — by mid- and late-encoder, MLP updates are roughly uniform across all tokens, consistent with the network preparing a globally pooled embedding. SAM, trained for dense prediction, retains spatial selectivity throughout: foreground tokens consistently receive distinctive MLP updates, with the butterfly outline still clearly visible at block 23. 

As expected from this observation, residual-consistency MLP fails catastrophically when applied to PE backbones, but performs surprisingly well on segmentation models.

\subsection{Imagenet Results} \label{sec:imagenet_res}
Based on the observation in the previous section, we retain the Stripe-Sort Attention mechanism while replacing the MLP compression approach with a more traditional method.token merging used in \cite{bolya2022token}.
\begin{figure}[t]
    \centering
    \includegraphics[width=\textwidth]{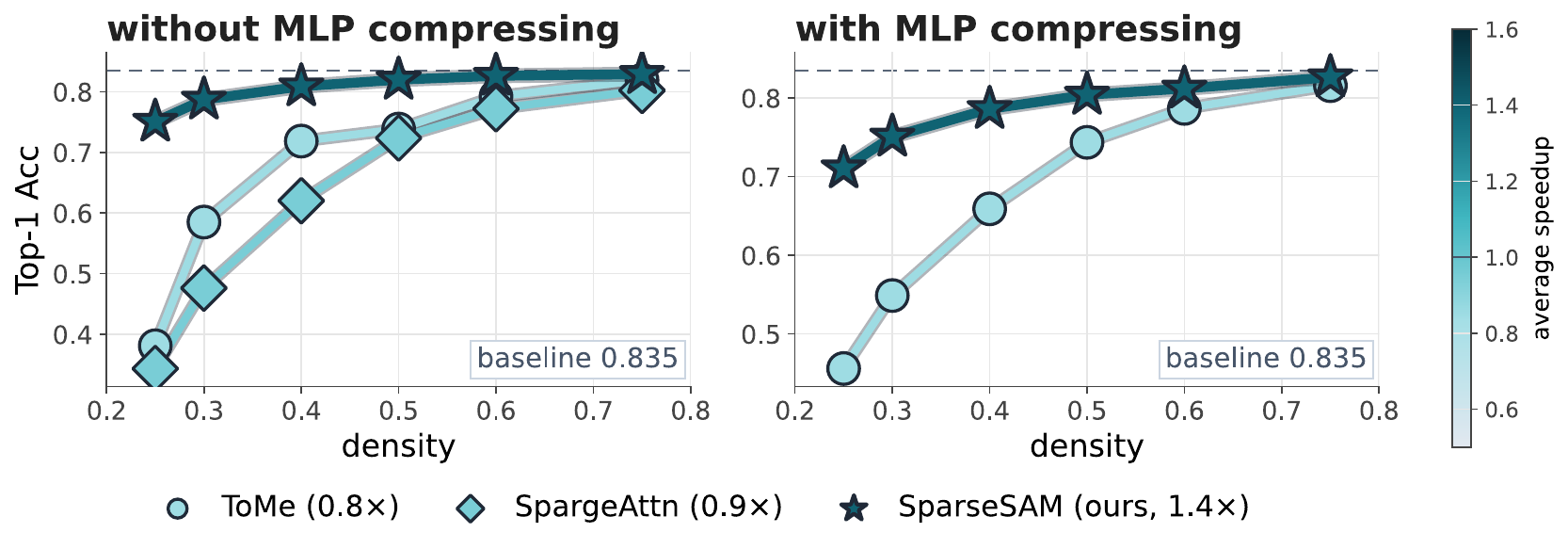}
    \caption{Results on ImageNet\cite{deng2009imagenet}}
    \label{fig:imagenet}
\end{figure}
Figure \ref{fig:imagenet} compares Top-1 accuracy across different sparsity densities under two settings: without MLP compression (left) and with MLP compression (right). In both cases, SparseSAM consistently outperforms token-merging baselines (ToMe) and sparse attention (SpargeAttn) across all density levels, maintaining performance close to or above the dense baseline (0.835). Without MLP compression, SparseSAM remains highly stable even at low densities (0.25–0.4), where competing methods show noticeable degradation. As density increases, all methods converge toward the baseline, but SparseSAM consistently stays ahead. When MLP compression is enabled, SparseSAM retains nearly identical accuracy trends while achieving higher efficiency (1.4× speedup). Notably, even under stronger compression, it preserves robustness across all density regimes, whereas baselines remain significantly below the dense reference, especially at low densities.

Overall, the results highlight that SparseSAM 's stripe-sort attention mechanism maintains a better accuracy-efficiency trade-off, with minimal degradation under joint attention and MLP compression.

\subsection{More segment output comparisons}
\begin{figure*}[!htb]
  \centering
  \includegraphics[width=0.9\textwidth]{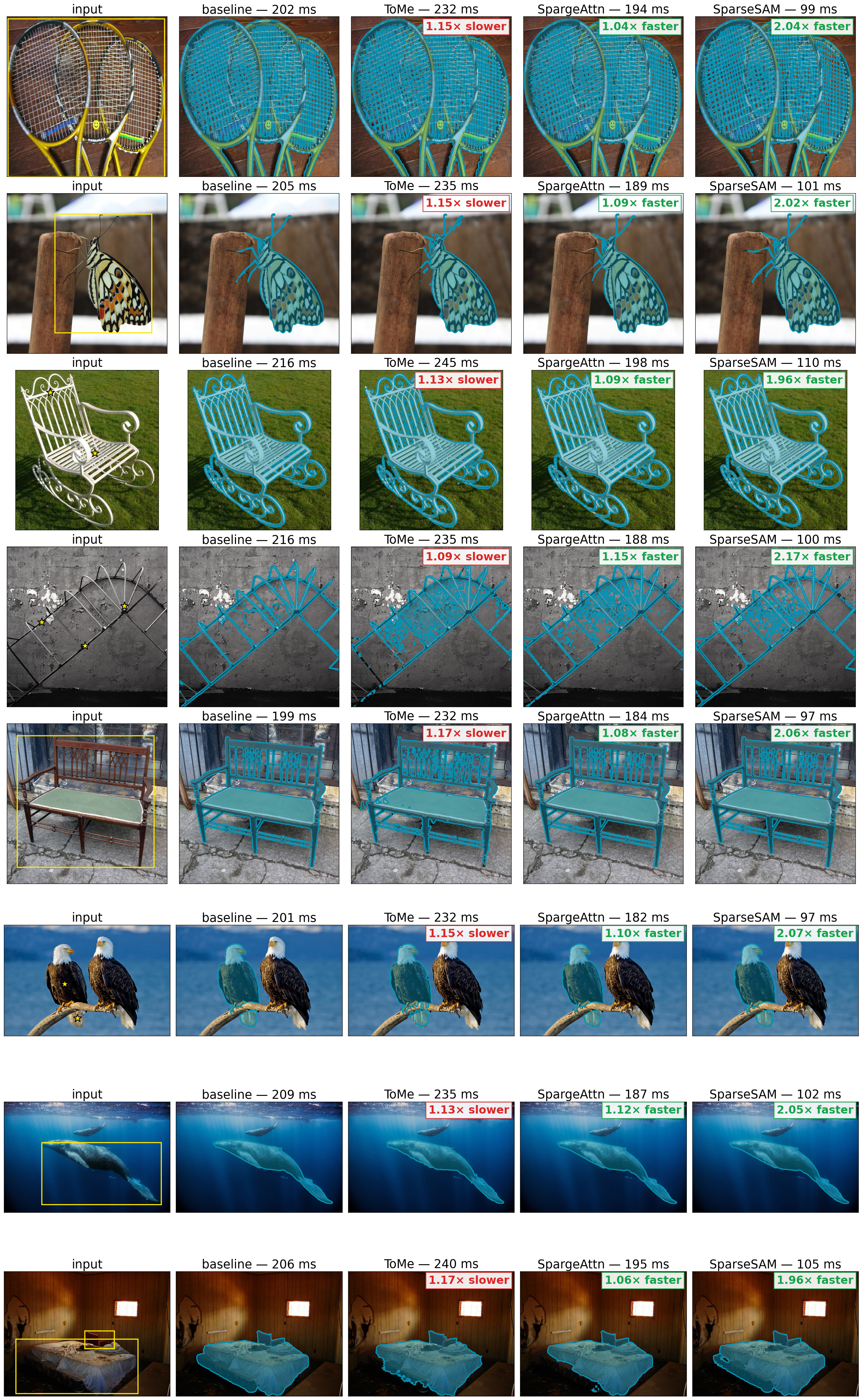}
  \caption{More output visualization}
  \label{fig:appendix-seg-examples}
\end{figure*}
\clearpage

\end{document}